\documentclass[preprint,12pt]{elsarticle}

\usepackage{amssymb}
\usepackage{amsmath}

\usepackage{algorithm}
\usepackage{algorithmicx}
\usepackage{algpseudocode}
\usepackage{amsfonts}
\usepackage{booktabs}
\usepackage{color}	
\usepackage[table]{xcolor}
\usepackage[a4paper]{geometry}
\usepackage{graphicx}
\usepackage{hyperref}
\usepackage{multirow}
\hypersetup{
	colorlinks=true,
	linkcolor=blue,
	filecolor=magenta,      
	urlcolor=cyan,
	pdftitle={Overleaf Example},
	pdfpagemode=FullScreen,
}

\usepackage{nomencl}
\usepackage{subcaption}
\usepackage{svg}
\usepackage{tabularray}
\usepackage{wrapfig}

\DeclareMathOperator*{\argmin}{\mathrm{argmin}}
\DeclareMathOperator*{\argmax}{\mathrm{argmax}}

\journal{Applied Mathematical Modelling}

\begin{document}
	\sloppy
	
	\begin{frontmatter}
		
		
		
		\title{Multi-fidelity Bayesian Data-Driven Design of Energy Absorbing Spinodoid Cellular Structures}

		
		\author[Delft-EEMCS]{Leo Guo} 
		
		\author[QMUL]{Hirak Kansara} 
		
		\author[QMUL]{Siamak F. Khosroshahi} 
		
		\author[Delft-EEMCS]{GuoQi Zhang\corref{cor1}} 
		\ead{g.q.zhang@tudelft.nl}
		
		\author[QMUL]{Wei Tan\corref{cor1}} 
		\ead{wei.tan@qmul.ac.uk}
        
		\cortext[cor1]{Corresponding author.}
		
			
			\affiliation[Delft-EEMCS]{organization={Department of Electronic Components, Technology and Materials, Delft University of Technology},
			addressline={Mekelweg 4}, 
			city={Delft},
			postcode={2628CM},
			country={Netherlands}}
			
		\affiliation[QMUL]{organization={School of Engineering and Materials Science, Queen Mary University of London},
			addressline={Mile End Road}, 
			city={London},
			postcode={E1 4NS},
			country={United Kingdom}}
		
		\begin{abstract}
			Finite element (FE) simulations of structures and materials are getting increasingly more accurate, but also more computationally expensive as a collateral result. This development happens in parallel with a growing demand of data-driven design. To reconcile the two, a robust and data-efficient optimization method called Bayesian optimization (BO) has been previously established as a technique to optimize expensive objective functions. In parallel, the mesh width of an FE model can be exploited to evaluate an objective at a lower or higher fidelity (cost \& accuracy) level. The multi-fidelity setting applied to BO, called multi-fidelity BO (MFBO), has also seen previous success. However, BO and MFBO have not seen a direct comparison with when faced with with a real-life engineering problem, such as metamaterial design for deformation and absorption qualities. Moreover, sampling quality and assessing design parameter sensitivity is often an underrepresented part of data-driven design. This paper aims to address these shortcomings by employing Sobol' samples with variance-based sensitivity analysis in order to reduce design problem complexity. Furthermore, this work describes, implements, applies and compares the performance BO with that MFBO when maximizing the energy absorption (EA) problem of spinodoid cellular structures is concerned. The findings show that MFBO is an effective way to maximize the EA of a spinodoid structure and is able to outperform BO by up to 11\% across various hyperparameter settings. The results, which are made open-source, serve to support the utility of multi-fidelity techniques across expensive data-driven design problems.
		\end{abstract}
		
		
		
		
	\end{frontmatter}
	
	\section{Introduction}
	
	Energy-absorbing structures, such as crumple zones, are essential in applications like vehicle safety, where their primary function is to convert kinetic energy into plastic deformation to protect occupants. By minimizing the forces experienced during collisions, these structures play a crucial role in the crashworthiness of vehicles, a key consideration in design and manufacturing. Traditionally, materials like steel and aluminum have been used in these applications. With advancements in 3D metal printing \cite{wu2023additively}, this process of additive manufacturing allows for the creation of complex prototypes that enhance energy absorption efficiency. More recently, however, there has been a shift towards using advanced composite materials and architected metamaterials, such as cellular structures \cite{harris2021impact} and shell-based structures \cite{han2015new,guell2019ultrahigh}, that offer superior strength-to-weight ratios and improved energy absorption.
	
	With the development of materials science, mechanical metamaterials have emerged as a promising solution for lightweight and efficient energy absorption (EA). Architected materials, including periodic structures like the aforementioned cellular solids, offer unique mechanical advantages but are often limited by stress concentrations that lead to premature failure. Innovations such as triply periodic minimal surfaces (TPMS) \cite{al2019multifunctional} introduced doubly curved, non-self-intersecting structures that distribute stress more evenly. However, while TPMS provides higher material stiffness, the aperiodic spinodoid metamaterials \cite{cahn1961spinodal} are more robust against imperfections in fabrication.
	
	Optimization of these complex energy-absorbing structures may require a strategic approach to balance multiple conflicting objectives, such as maximizing EA while minimizing peak forces (PF) experienced during impact. Conventional design methods often rely on trial-and-error or topology optimization \cite{bendsoe2013topology,zhang2021mechanical,guo2024inverse}. Meanwhile, topology optimization iteratively redistributes material within a design domain but can produce geometries with sharp angles and irregularities that complicate fabrication. With both of these traditional methods having major drawbacks, data-driven design is a strong contender to address this type of optimization problem.
	
	One of the cornerstones of the data-driven design process is design space sampling. Popular examples include grid sampling \cite{yu2018mechanical}, uniform random sampling and Latin hypercube sampling. While these methods have proven to be effective in some use cases, Sobol' sampling has proven to be more effective in gaining insight when data is scarce due to an expensive objective: by using Sobol' samples, it is possible to perform variance-based sensitivity analysis of the objective with respect to the design parameters.
	
\begin{figure}[H]
		\centering
		\includesvg[width=\linewidth]{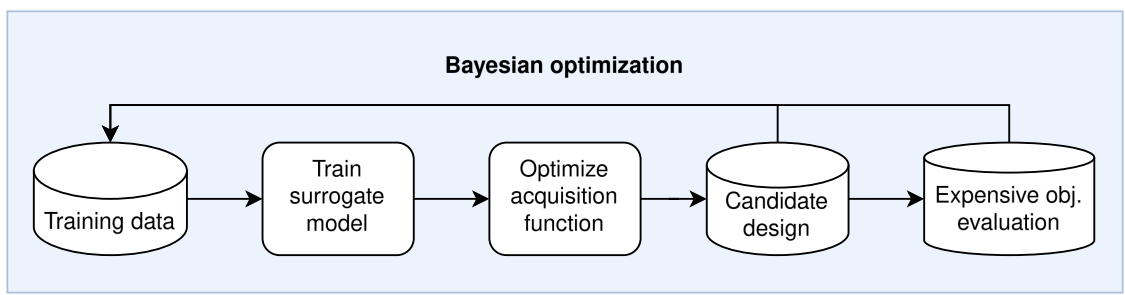}
		\caption{Flowchart diagram of BO.}
		\label{fig:bo-diagram}
	\end{figure}

	However, sampling alone is not sufficient to provide a robust objective optimization scheme. In literature, data-driven optimization approaches leverage machine learning models as surrogates for costly simulations. Among them, Bayesian optimization (BO) with Gaussian process regression has gained traction for optimizing meta-material structure designs with high computational cost \cite{bessa2019bayesian, shin2021spiderweb, kuszczak2023bayesian}, including spinodoid structures \cite{rassloff2025inverse}. BO is a popular data-efficient objective optimization methodology. It aims to leverage the proven regression capabilities of Bayesian machine learning methods to assess the likelihood of the optimum location. This leads to a delicate balance between exploring new regions of high uncertainty where the black box function may have a global (or at least a better local) optimum, and exploiting regions where an optimum is already expected to be present. It trains a surrogate model on a set of training data, populated by evaluated design experiments, and subsequently optimizes a proxy belief model based on the surrogate regression model -- also called an acquisition function. The design that is found through this process is then evaluated and augmented to the original training data, after which the cycle starts anew. See Figure \ref{fig:bo-diagram} for a schematic overview of BO.
	
	BO can be flexibly adapted into settings with data that adhere to particular input or output structures.	Multi-fidelity BO (MFBO), which relies on surrogate models that handle the cost-accuracy trade-off between low- and high-fidelity data, allows for accelerated convergence by leveraging inexpensive approximations alongside selective high-fidelity evaluations. This approach enables more comprehensive exploration of the design space while maintaining the accuracy needed for realistic applications. Numerous successful mechanical engineering design optimizations have been performed by using MFBO. These cases include mathematical and physical phenomena with complex behavior such as inverted pendulums \cite{marco2017virtual}, fractional advection-dispersion equations \cite{pang2017discovering}, haemodynamics \cite{perdikaris2016model} and non-linear state-space models \cite{imani2019mfbo}. Also, engineering design processes have been developed with MFBO, including the design of antennas \cite{wu2020multistage}, fixed-wing drones \cite{charayron2023towards}, helicopter blades \cite{ariyarit2018hybrid}, aerofoils \cite{tyan2015improving,kou2019multi,peng2023multi}, flip-chip packages \cite{tran2020smf}, horizontal road alignment \cite{aziz2020multi}, and analog circuits \cite{zhang2019efficient}.
	
	The multi-fidelity techniques has proven particularly useful when the objective functions are provided by finite element (FE) simulations, where the fidelity trade-off is intuitively encoded by the mesh size \cite{li2020multi,cupertino2023centimeter}. By incorporating elastic-plastic material behaviors into quasi-static FE compression simulations, we capture complex structural responses under impact. The positive track record of MFBO applications renders the EA optimization problem a suitable problem to tackle with a multi-fidelity optimization methodology.
	
	
	This work presents a novel framework that addresses two key gaps in current optimization practice: the lack of direct comparisons between MFBO and its single-fidelity counterpart, and the underexplored role of design space sampling in sensitivity analysis. We introduce a two-stage approach that first applies Sobol' and Saltelli sampling to perform global, variance-based sensitivity analysis of the EA objective across fidelity levels—offering new insight into how parameter importance shifts with model accuracy. We then apply both BO and MFBO to optimize spinodoid topologies for EA, enabling a rigorous, side-by-side evaluation of optimization performance. This study is among the first to demonstrate the practical benefits and trade-offs of MFBO in complex, physics-driven design spaces. To support transparency and reproducibility, all code and data are made openly available via GitHub. The manuscript concludes with a critical analysis of MFBO's applicability to safety-critical materials design, outlining future directions for advancing multi-fidelity optimization strategies.
	
	\section{Data-driven methodology}
	
	\subsection{Sobol' sensitivity analysis}
	\label{sec:ssa}
	
	Let $f:[0,1]^D\to\mathbb{R}$ be an objective function of the design parameters $x_1,x_2,\ldots,x_D$. In order to gain an initial insight on the response of $f$ across the (scaled) design space, it is essential to know how much $f$ varies by perturbing the design parameters. This is precisely the goal of sensitivity analysis, which is a common technique used in practice when phenomena involving expensive or scarce data are concerned, such as hydrology \cite{lenhart2002comparison} and financial decision making \cite{baucells2013invariant}. Indeed, a review of various methods and applications identifies sensitivity analysis as part of the best practices to ensure modelling quality in general \cite{borgonovo2016sensitivity}.
	
	Originally introduced by Sobol' \cite{sobol2001global}, the Sobol' sensitivity analysis (SSA) approach treats the objective and design parameters $f(X_1,\ldots,X_D)=Y$ as fully stochastic, with $X_1,\ldots X_D,Y$ being modeled as random variables. Next, let $I=\{i_1,\ldots,i_d\}\subset\{1,\ldots,D\}$ be a subset of size $d$, also called a $d$-set. Define $\textbf{X}_{I}:=(X_{i_1},\ldots,X_{i_d})^\top$. 
	By the law of total variance,
	$$\text{Var}(Y)=\mathbb{E}(\text{Var}(Y|\textbf{X}_{I}))+\text{Var}(\mathbb{E}(Y|\textbf{X}_{I})),$$
	which splits the total, unconditional variance of the dependent variable $Y$ into its unexplained and explained components with respect to the selected design parameters according to the indices in $I$.
	
	By dividing the explained variance component $\text{Var}(\mathbb{E}(Y|\textbf{X}_{I}))$ by the total variance of $Y$, one will obtain a normalized variance metric, also called the Sobol' sensitivity index \cite{opgenoord2016variance, marelli2019uqlab} with respect to $\textbf{X}_{I}$:
	\begin{equation}
		S_{I}:=\frac{\text{Var}(\mathbb{E}(Y|\textbf{X}_{I}))}{\text{Var}(Y)}.
		\label{eq:sensitivity-index-set-i}
	\end{equation}
	The special univariate case where $I=\{i\}$ for some $i\in\{1,\ldots,D\}$,
	\begin{equation}
		S_i:=\frac{\text{Var}(\mathbb{E}(Y|X_i))}{\text{Var}(Y)},
		\label{eq:Si}
	\end{equation}
	is called the first order sensitivity index with respect to design parameter $X_i$, and is of particular interest.
	
	Dual to the first order index $S_i$ is the concept of total effect index with respect to $X_i$, defined as the ratio between the \textit{unexplained} variance of all design variables \textit{except} $X_i$, and the total variance:
	\begin{equation}
		S_{T_i}:=\frac{\mathbb{E}(\text{Var}(Y|\textbf{X}_{\sim i}))}{\text{Var}(Y)}=1-\frac{\text{Var}(\mathbb{E}(Y|\textbf{X}_{\sim i}))}{\text{Var}(Y)},
		\label{eq:STi}
	\end{equation}
	where $\sim i:=\{1,\ldots,D\}\backslash\{i\}$.
	
	Calculating the required statistical moments for $S_i$ and $S_{T_i}$ generally implies the calculation of multiple intractable integrals, and is therefore almost always approximated in literature. A common approximations to evaluate the expressions in Equation \eqref{eq:Si} and Equation \eqref{eq:STi} are given by Saltelli et al. \cite{saltelli2010variance}, as briefly described below.
	
	Let $\textbf{x}_1,\textbf{x}_2,\ldots,\textbf{x}_N\in[0,1]^{2D}$ be the first $N$ terms a $2D$-Sobol' sequence. To be explicit, let $\textbf{x}_j:=(x_{j,1},x_{j,2},\ldots,x_{j,2D})^\top$ be the components of $\textbf{x}_j$ for any $j\in\{1,\ldots,N\}$. Next, define 
	\begin{align}
		\textbf{A}:=(\textbf{a}_1^\top,\ldots,\textbf{a}_N^\top)^\top\text{ where }\textbf{a}_j&:=(x_{j,1},\ldots,x_{j,D})^\top,\\
		\textbf{B}:=(\textbf{b}_1^\top,\ldots,\textbf{b}_N^\top)^\top\text{ where }\textbf{b}_j&:=(x_{j,D+1},\ldots,x_{j,2D})^\top.
	\end{align}
	
	Next, for any $i\in\{1,\ldots,D\}$, define the matrix $\textbf{A}_{\mathbf{B}}^{(i)}$ as the $N\times D$ matrix containing coordinate-wise cross-combinations of rows that populate $\textbf{A}$ and $\textbf{B}$:
	
	\begin{equation}
		(\textbf{A}_{\mathbf{B}}^{(i)})_{j,k}:=\left\{
		\begin{array}{cc}
			a_{j,k},&k\ne i,\\
			b_{j,k},&k=i,
		\end{array}
		\right.
		\label{eq:A-B-mix-construction}
	\end{equation}
	where $j\in\{1,\ldots,N\}$, $k\in\{1,\ldots,D\}$ and the component notations are derived from $\textbf{a}_j=:(a_{j,1},\ldots,a_{j,D})^\top$ and $\textbf{b}_j=:(b_{j,1},\ldots,b_{j,D})^\top$. See Figure \ref{fig:ssa-diagram-general} for a graphic illustration on how to construct $\textbf{A}_{\textbf{B}}^{(i)}$.
	
	\begin{figure}[H]
		\centering
		\includesvg[width=.9\linewidth]{img/diagrams/ssa_diagram_general.drawio.svg}
		\caption{A schematic construction of $\textbf{A}_{\textbf{B}}^{(i)}$ given $\textbf{A}$ and $\textbf{B}$.}
		\label{fig:ssa-diagram-general}
	\end{figure}
	
	Finally, let $\textbf{A}_{\mathbf{B}}^{(i)}:=(\textbf{c}_1^{(i)\top},\ldots,\textbf{c}_{N}^{(i)\top})^\top$ be decomposed into rows. Then, the following approximations hold:
	\begin{align}
		\mathrm{Var}(\mathbb{E}(Y|X_i))&\approx \frac1N\sum_{j=1}^Nf(\textbf{b}_j)(f(\textbf{c}_j^{(i)})-f(\textbf{a}_j)),\label{eq:Si-unnormalized}\\
		\mathbb{E}(\mathrm{Var}(Y|\textbf{X}_{\sim i}))&\approx\frac1{2N}\sum_{j=1}^N(f(\textbf{a}_j)-f(\textbf{c}_j^{(i)}))^2.\label{eq:STi-unnormalized}
	\end{align}
	Moreover, the statistical confidence of these approximations can be numerically tested by means of bootstrapping \cite{marelli2019uqlab,efron1992bootstrap}.
	
\begin{figure}[H]
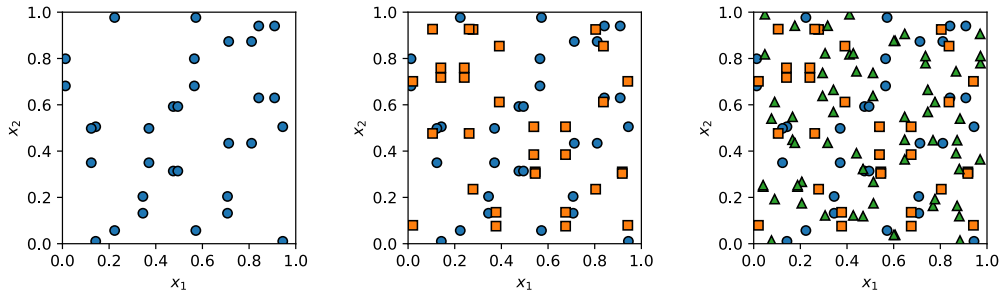

		\centering
		\begin{subfigure}{.3\textwidth}
			\includesvg[width=\textwidth]{img/diagrams/Saltelli_N32_add.svg}
		\end{subfigure}
		\begin{subfigure}{.3\textwidth}
			\includesvg[width=\textwidth]{img/diagrams/Saltelli_N64_add.svg}
		\end{subfigure}
		\begin{subfigure}{.3\textwidth}
			\includesvg[width=\textwidth]{img/diagrams/Saltelli_N128_add.svg}
		\end{subfigure}
		\caption{Saltelli sampling on a unit square with 32, 64 and 128 subsequent samples (different colors and shapes). These samples are respectively based on 8, 16 and 32 subsequent Sobol' samples in 4D space.}
		\label{fig:sampling-saltelli}
	\end{figure}

	An aggregate of all design samples necessary to perform SSA, i.e. $\textbf{A},\textbf{B}$ and $\textbf{A}_{\textbf{B}}^{(i)}$ for all $i\in\{1,\ldots D\}$, is called a collection of Saltelli samples, where it should be noted that the quantity of these samples is always a multiple of $D+2$. An example of two-dimensional Saltelli sampling is given in Figure \ref{fig:sampling-saltelli}.
	
	From Figure \ref{fig:sampling-saltelli}, it can be seen that the sequentially dependent distribution of the Saltelli samples fills the space. This makes Saltelli sampling the preferred method to access samples, not only to perform sensitivity analysis, but also to be used as a data set to build a regression model on, particularly as a central step in the Bayesian optimization algorithm.
	
	\subsection{Bayesian optimization}
	\label{sec:gp}
	
	In literature, a wide variety of regression models exist to model the relationship between $\textbf{X}$ and $\textbf{y}$. In terms of complexity, models range from linear ones such as ordinary least squares, ridge \cite{hoerl1970ridge} or lasso \cite{tibshirani1996regression} regression to the highly non-linear artificial neural network (ANN) models. Among these, some regression models have been successfully employed from which posterior information can be extracted in order to suggest new data in the Bayesian optimization. Examples include random forest regression \cite{hutter2011sequential}, tree-structured Parzen estimators \cite{bergstra2011algorithms} and Bayesian ANNs (BNNs) \cite{snoek2015scalable}. However, as the prominent reviewers of Bayesian optimization methods have iterated \cite{shahriari2015taking,garnett2023bayesian}, the Gaussian process model is the most popular type of model used in practice to perform regression within the Bayesian optimization context.
	
	\subsubsection{Gaussian process regression}
	
	Given a design of experiments $\textbf{D}$ with $N$ experiments, it is assumed that $\textbf{X}$ is an independent variable with respect to an explanatory model, while $\textbf{y}$ is a dependent variable.
	
	In the context of Bayesian statistics, it is assumed that any given set of evaluations of $f$ is the realization of a jointly distributed random variable. Specifically, $f$ could be modelled as a Gaussian process (GP) with zero\footnote{This is a simplifying assumption in concordance with $f$ being standardized. However, it is not strictly necessary \cite{bishop2006pattern, williams2006gaussian}.} prior mean. Concretely, for any $\textbf{x}\in[0,1]^D$, it is assumed that $f(\textbf{x})$ is a random variable with $\mathbb{E}(f(\textbf{x}))=0$ and prior covariance function or kernel $\kappa$; for any $\textbf{u},\textbf{v}\in[0,1]^D$, the defining property is as such:
	\begin{equation}
		\kappa(\textbf{u},\textbf{v}):=\text{Cov}(f(\textbf{u}), f(\textbf{v})).
	\end{equation}
	
	Let $\theta_1,\ldots,\theta_{T_\kappa}$ be the GP model parameters, collected into a model parameter vector $\boldsymbol{\theta}=(\theta_1,\ldots,\theta_{T_\kappa})$. Since $\boldsymbol{\theta}$ contains precisely the model parameter coordinates on which $\kappa$ depends, the notation $\kappa(\textbf{u},\textbf{v})=\kappa_{\boldsymbol{\theta}}(\textbf{u},\textbf{v})$ arises.
	
	A popular type of prior covariance function is the radial basis kernel (RBF) kernel \cite{williams2006gaussian, schulz2018tutorial, liu2018gaussian}, given by:
	\begin{equation}
		\kappa_{\text{RBF},\boldsymbol{\theta}_{\text{RBF}}}(\textbf{u},\textbf{v}):=c\cdot\exp\left(-\frac{\|\textbf{u}-\textbf{v}\|^2}{2\lambda^2}\right)+s^2\delta(\textbf{u}-\textbf{v}),
		\label{eq:RBF-simple}
	\end{equation}
	where $c$ is called the amplitude parameter, $\lambda$ is the length scale parameter and $s^2$ is the aleatoric noise variance parameter. The noise variance, as the name suggests, only affects the scenario in which the covariance is calculated for $\textbf{u}=\textbf{v}$ as indicated by the Dirac-$\delta$ function. In summary, the GP model parameters associated with the RBF kernel are $\boldsymbol{\theta}_{\text{RBF}}:=(c,\lambda,s^2)^\top$.
	
	As dependent variables in a Bayesian context, it is assumed that $\textbf{y}$ is a realization of a random vector $\textbf{Y}$, and that any model parameter vector $\boldsymbol{\theta}$ is a realization of a random vector $\boldsymbol{\Theta}$. 
	The Gaussian process assumption implies that $\textbf{Y}$ follows a multivariate normal distribution when a model parameter vector $\boldsymbol{\theta}$ is established. In statistical terms: the prior distribution of $\textbf{Y}$ conditioned on $\boldsymbol{\Theta}=\boldsymbol{\theta}$ is given by
	\begin{equation}
		\textbf{Y}|(\boldsymbol{\Theta}=\boldsymbol{\theta})\sim\mathcal{N}(\textbf{0},\textbf{K}_{\boldsymbol{\theta}}),
		\label{gp-prior}
	\end{equation}
	with covariance matrix $\textbf{K}_{\boldsymbol{\theta}}=\textbf{K}_{\boldsymbol{\theta}}(\textbf{X}):=(\kappa_{\boldsymbol{\theta}}(\textbf{x}_i,\textbf{x}_j))_{i,j=1,\ldots,N}$.
	
	For any given $\textbf{x}\in[0,1]^D$, it is now of interest to find the distribution of $f(\textbf{x})$ given $\textbf{y}$ and $\boldsymbol{\theta}$, which is also called the (parameterized) posterior distribution. Here lies the crux of assuming $f$ to be  a GP, since $(\textbf{Y}^\top, f(\textbf{x}))^\top$ is another multivariate normal random variable. Then, following some mathematical manipulations \cite{williams2006gaussian, holt2023essential}, it can be shown that the parameterized posterior is also multivariate normal:
	\begin{equation}
		f(\textbf{x})|(\textbf{Y}=\textbf{y},\boldsymbol{\Theta}=\boldsymbol{\theta})\sim\mathcal{N}(\mu_{\boldsymbol{\theta}}(\textbf{x}),\sigma_{\boldsymbol{\theta}}^2(\textbf{x})),
		\label{eq:intermediate-ppd}
	\end{equation}
	where
	\begin{align}
		\mu_{\boldsymbol{\theta}}(\textbf{x})&:=\kappa_{\boldsymbol{\theta}}(\textbf{x},\textbf{X})^\top\textbf{K}_{\boldsymbol{\theta}}^{-1}\textbf{y},
		\label{gp-ppd-mean}\\
		\sigma_{\boldsymbol{\theta}}^2(\textbf{x})&:=\kappa_{\boldsymbol{\theta}}(\textbf{x},\textbf{x})-\kappa_{\boldsymbol{\theta}}(\textbf{x},\textbf{X})^\top \textbf{K}_{\boldsymbol{\theta}}^{-1}\kappa_{\boldsymbol{\theta}}(\textbf{x},\textbf{X}),
		\label{gp-ppd-var}
	\end{align}
	with $\kappa_{\boldsymbol{\theta}}(\textbf{x},\textbf{X}):=(\kappa_{\boldsymbol{\theta}}(\textbf{x},\textbf{x}_1),\ldots,\kappa_{\boldsymbol{\theta}}(\textbf{x},\textbf{x}_N))^\top$ as a column vector of length $N$. It is also convenient to denote $\mu_{\boldsymbol{\theta}}(\textbf{Z}):=(\mu_{\boldsymbol{\theta}}(\textbf{z}_1),\ldots,\mu_{\boldsymbol{\theta}}(\textbf{z}_P))^\top$ as a column vector of length $P$ whenever $\textbf{Z}=(\textbf{z}_1^\top,\ldots,\textbf{z}_P^\top)^\top$ is a column of $P$ design rows. The same notation is introduced for $\sigma^2_{\boldsymbol{\theta}}$.
	
	While Equation \eqref{eq:intermediate-ppd} provides a closed form description of the (parameterized) posterior distribution of $f(\textbf{x})|(\textbf{Y}=\textbf{y},\boldsymbol{\Theta}=\boldsymbol{\theta})$, it is desirable to obtain the distribution of $f(\textbf{x})|(\textbf{Y}=\textbf{y})$, also called the posterior predictive distribution (PPD), independent of the model parameter vector $\boldsymbol{\theta}$. 
	
	In practice, the PPD cannot be calculated analytically \cite{williams2006gaussian}. In order to handle this, the dependency on the model parameter vector $\boldsymbol{\theta}$ of the intermediate posterior predictive distribution is removed by approximating it with a point estimate. 
	
	One choice for such estimate, in the context of Gaussian process modelling, is the maximum (log) likelihood estimate (MLE)  \cite{williams2006gaussian}. By defining the model parameter likelihood function as
	\begin{equation}
		\ell(\boldsymbol{\theta;\textbf{D}}):=p(\textbf{Y}|\boldsymbol{\Theta}=\boldsymbol{\theta})[\textbf{y}],
		\label{model-parameter-likelihood}
	\end{equation}
	the MLE is given by
	\begin{equation}
		\hat{\boldsymbol{\theta}}_{\text{MLE}}=\argmin_{\boldsymbol{\theta}\in{\mathcal{T}_\kappa}}\ln(\det(\textbf{K}_{\boldsymbol{\theta}}(\textbf{X})))+\textbf{y}^\top \textbf{K}^{-1}_{\boldsymbol{\theta}}(\textbf{X})\textbf{y},
		\label{eq:mle}
	\end{equation}
	where $\mathcal{T}_\kappa:=P_1\times P_2\times\cdots \times P_{T_\kappa}\subset\mathbb{R}^{T_\kappa}$ be the $T_\kappa$-dimensional domain of model parameter vectors $\boldsymbol{\theta}$ and where $P_t\subset\mathbb{R}$ is defined to be the domain of all permissible values of $\theta_t$, $t\in\{1,\ldots T_k\}$.
	
	Because $\ell$ equals the probability density function of a normal distribution as a function of $\textbf{y}$, this likelihood function can also be referred to as a Gaussian likelihood function.
	
	Some calculation yields the following approximation for the PPD density:
	\begin{equation}
		p(f(\textbf{x})|\textbf{Y}=\textbf{y})[y]\approx p(f(\textbf{x})|\textbf{Y}=\textbf{y},\boldsymbol{\Theta}=\hat{\boldsymbol{\theta}}_{\text{MLE}})[y]=\phi_{\hat{\mu}(\textbf{x}),\hat{\sigma}^2(\textbf{x})}(y)=:\hat{\phi}(y;\textbf{x},\textbf{y},\kappa),
		\label{ppd-density-approx}
	\end{equation}
	where 
	$$\phi_{\mu,\sigma^2}(z):=\frac{1}{\sigma\sqrt{2\pi}}\exp\left(-\frac12\left(\frac{z-\mu}{\sigma}\right)^2\right)$$
	denotes the normal probability density function with mean $\mu$ and variance $\sigma^2$. Furthermore, the notations
	\begin{equation}
		\hat{\mu}:=\mu_{\hat{\boldsymbol{\theta}}_{\text{MLE}}},\quad\hat{\sigma}^2:=\sigma^2_{\hat{\boldsymbol{\theta}}_{\text{MLE}}}
		\label{ppd-MLE}
	\end{equation}
	are used for the optimized parameterized posterior parameters in Equation \eqref{gp-ppd-mean} and Equation \eqref{gp-ppd-var}. 
	
	Commonly, it is assumed that the model parameter likelihood $\ell$, Equation \eqref{model-parameter-likelihood}, is differentiable with respect to the model parameter vector $\boldsymbol{\theta}$. Then, due to the differentiable nature of the optimization problem in Equation \eqref{eq:mle}, it is most often solved by using gradient-based methods such as L-BFGS or Adam.
	
	It is possible to shape the information that $\hat{\phi}^{(i)}$ provides into a design selection policy $\Pi$ that suggests a candidate $\textbf{x}^{(i)}\in[0,1]^D$ for objective evaluation. In short, this is achieved by means of a so-called acquisition function or infill function $\alpha:[0,1]^D\to\mathbb{R}$, which intuitively assigns a measure of belief $\alpha(\textbf{x};\hat{\phi}^{(i)})$ to the design $\textbf{x}\in[0,1]^D$ that $f(\textbf{x})$ is able to achieve a better optimum than $f(\textbf{z})$ for any design row $\textbf{z}^\top$ in $\textbf{X}^{(i-1)}$.
	
	As $\alpha$ expresses a level of belief across the design parameter space, design selection takes place by locating such $\textbf{x}\in[0,1]^D$ at which $\alpha$ is maximized: see Equation \ref{eq:acq-opt}.
	
	\begin{equation}
		\textbf{x}^{(i)}:=\argmax_{\textbf{x}\in[0,1]^D}\alpha(\textbf{x};\hat{\phi}^{(i)})
		\label{eq:acq-opt}
	\end{equation}
	
	An extensive set of acquisitions have been developed since the inception of the Bayesian optimization scheme. Popular examples of acquisition functions $\alpha$ are the expected improvement (EI) acquisition \cite{jones1998efficient} and the upper/lower confidence bound (UCB) acquisition \cite{srinivas2009gaussian}. See Equation \eqref{eq:ei} and Equation \eqref{eq:ucb} respectively.
	
	\begin{align}
		\alpha_{\text{EI}}(\textbf{x};\hat{\phi}^{(i)})&:=\hat{\sigma}^{(i)}(\textbf{x})(z^{(i)}(\textbf{x})\hat{\Phi}(z^{(i)}(\textbf{x}))+\hat{\phi}(z^{(i)}(\textbf{x})))
		\label{eq:ei}\\
		\alpha_{\text{UCB}}(\textbf{x};\hat{\phi}^{(i)},\beta)&:=\hat{\mu}^{(i)}(\textbf{x})+\beta\hat{\sigma}^{(i)}(\textbf{x})
		\label{eq:ucb}
	\end{align}
	
	In the UCB acquisition function, $\beta$ is a hyperparameter.
	
	In practice, a logarithmic variant of the EI acquisition function (LogEI) is used due to numerical stability during optimization \cite{ament2023unexpected}. The derivations and intuition behind these formulations can be referred to in \cite{garnett2023bayesian,shahriari2015taking}.
	
	The BO schematic introduced in Figure \ref{fig:bo-diagram} is summarized in Algorithm \ref{algo:bo} when GPR is used as a surrogate modelling methodology.
	
	\begin{algorithm}[H]
		\caption{Bayesian optimization with Gaussian process regression (BO)}
		\begin{algorithmic}[1]
			\Require{Design of training experiments $\textbf{D}^{(0)}$, covariance function $\kappa$, acquisition function $\alpha$, number of iterations $I$}
			\For{$i=1,\ldots,I$}
			\State\small$\hat{\boldsymbol{\theta}}_{\text{MLE}}^{(i)}\gets\argmin_{\boldsymbol{\theta}\in\mathcal{T}_\kappa}\ln(\det(\textbf{K}_{\boldsymbol{\theta}}(\textbf{X}^{(i-1)})))+\textbf{y}^{(i-1)\top} \textbf{K}^{-1}_{\boldsymbol{\theta}}(\textbf{X}^{(i-1)}) \textbf{y}^{(i-1)}$
			\State $\hat{\phi}^{(i)}\gets\hat{\boldsymbol{\theta}}_{\text{MLE}}^{(i)}$
			\State $\textbf{x}^{(i)}\gets\argmax_{\textbf{x}\in[0,1]^D}\alpha(\textbf{x};\hat{\phi}^{(i)})$
			\State $y^{(i)}\gets f(\textbf{x}^{(i)})$
			\State $\textbf{D}^{(i)}\gets(\textbf{D}^{(i-1)},(\textbf{x}^{(i)\top},y^{(i)}))^\top$
			\EndFor
			\State $(\textbf{x}_{\text{rec}},y_{\text{rec}})\gets\text{Rec}(\textbf{D}^{(I)})$
			\\\Return $(\textbf{x}_{\text{rec}},y_{\text{rec}})$
		\end{algorithmic}
		\label{algo:bo}
	\end{algorithm}
	
	In step 8 of Algorithm \ref{algo:bo}, The Rec function returns the recommended design-objective pair that is found throughout the optimization process. This corresponds to the design that achieves the minimized objective throughout all of the iterations.
	
	\subsubsection{Multi-output Gaussian processes}
	\label{sec:mogp}
	
	Given the scarcity of data to construct a design of experiments $\textbf{D}$, a common solution is to acquire higher-throughput data, at a risk of such data having larger (aleatoric) uncertainty or inherent biases compared to $\textbf{D}$. The trade-off between these levels of fidelity gives rise to a data structure which consists of a large volume of low-cost, low-accuracy data and a limited volume of high-cost, high-accuracy data \cite{fernandez2016review}. Prioritizing efficiency over accuracy can be achieved by experimental \cite{forrester2010black, cupertino2023centimeter}, computational \cite{balabanov1998multifidelity, fidkowski2014quantifying} or analytical strategies \cite{forrester2007multi, le2014bayesian}. The assumption throughout this manuscript will be such that there are an $M$ number of such trade-offs, each of which is associated with a single-output objective function $f_m:[0,1]^D\to\mathbb{R}$, where $m\in\{1,\ldots,M\}$. The utilized convention is that $f_1$ has the lowest fidelity (i.e., cost and accuracy), and $f_M$ has the highest fidelity. The data structure of a multi-fidelity design of experiments will be generically written as a vertical concatenation $\boldsymbol{\mathcal{D}}:=(\textbf{D}_1,\ldots,\textbf{D}_M)^\top$ of $M$ DoEs, ordered by increasing levels of fidelity. Similar notation for the multi-fidelity input data $\boldsymbol{\mathcal{X}}:=(\textbf{X}_1,\ldots,\textbf{X}_M)^\top$ and multi-fidelity output data $\textbf{y}:=(\textbf{y}_1,\ldots,\textbf{y}_M)^\top$ are introduced, where $\textbf{D}_m:=(\textbf{X}_m,\textbf{y}_m)$ for $m\in\{1,\ldots,M\}$. It is furthermore assumed that $\textbf{X}_m$ consists of $N_m$ designs with $D$ design variables, and therefore has dimensions $N_m\times D$, while $\textbf{y}_m$ is correspondingly a column vector of length $N_m$ consisting of the (realized) values $f_m(\textbf{X}_m)$. 
	
	A single-output objective $f$ can be treated like a (single-output) GP (SOGP). This can be generalized to handling a multi-output objective $\textbf{f}:=(f_1,\ldots,f_M)^\top$. In the case of a SOGP model, the covariance kernel $\kappa$ is a real-valued function. A multi-output Gaussian process (MOGP) model with $M$ outputs generalizes $\kappa$ an $M\times M$ matrix-valued function: $\boldsymbol{\mathcal{K}}(\cdot,\cdot):=(\kappa_{i,j}(\cdot,\cdot))_{i,j=1}^M$. As with the SOGP case, the covariance $\boldsymbol{\mathcal{K}}$ is dependent on a vector of model parameters $\boldsymbol{\theta}$, which for the sake of notational clarity will be suppressed for the remainder of this subsection. Each component $\kappa_{i,j}$ is to be interpreted as the (single-output) covariance function between fidelities $i$ and $j$. For example, if $\textbf{u}$ is a design to be evaluated at fidelity $i$ and $\textbf{v}$ at fidelity $j$,
	\begin{equation}
		\kappa_{i,j}(\textbf{u},\textbf{v}):=\text{Cov}(f_i(\textbf{u}),f_j(\textbf{v})).
	\end{equation}
	
	Following similar reasoning as for SOGP models, the parameterized posterior distribution is an $M$-dimensional multivariate normal distribution,
	\begin{equation}
		\textbf{f}(\textbf{x})|(\textbf{Y}=\textbf{y},\boldsymbol{\Theta}=\boldsymbol{\theta})\sim\mathcal{N}(\boldsymbol{\mu}(\textbf{x}),\boldsymbol{\Sigma}(\textbf{x})),
	\end{equation} 
	which models each fidelity simultaneously. The parameterized posterior distribution comprises $\boldsymbol{\mu}(\textbf{x})=(\mu_1(\textbf{x}),\ldots,\mu_M(\textbf{x}))^\top$ as the posterior $M$-dimensional mean and $\boldsymbol{\Sigma}(\textbf{x})=(\sigma_{i,j}^2(\textbf{x}))_{i,j=1,\ldots,M}$ as the $M\times M$ posterior covariance matrix, whenever $\textbf{x}\in[0,1]^D$. The components of these statistical parameters are readily interpretable as such: $\mu_m(\textbf{x})$ is the mean and $\sigma_m^2(\textbf{x}):=\sigma_{m,m}^2(\textbf{x})$ is the variance at fidelity $m\in\{1,\ldots,M\}$.
	
	Despite the introduced complexity compared to SOGP, their expressions in terms of the prior covariance function $\boldsymbol{\mathcal{K}}$ are similar, cf. \eqref{gp-ppd-mean}, \eqref{gp-ppd-var}:
	\begin{align}
		\boldsymbol{\mu}(\textbf{x})&=\boldsymbol{\mathcal{K}}(\textbf{x},\boldsymbol{\mathcal{X}})^\top \textbf{K}^{-1}\textbf{y},\label{mogp-ppd-mean}\\
		\boldsymbol{\Sigma}(\textbf{x})&=\boldsymbol{\mathcal{K}}(\textbf{x},\textbf{x})-\boldsymbol{\mathcal{K}}(\textbf{x},\boldsymbol{\mathcal{X}})^\top \textbf{K}^{-1}\boldsymbol{\mathcal{K}}(\textbf{x},\boldsymbol{\mathcal{X}}).\label{mogp-ppd-var}
	\end{align}
	In the MOGP scenario, the covariance matrix
	\begin{equation}
		\textbf{K}:=\begin{pmatrix}
			\textbf{K}_{1,1}(\textbf{X}_1,\textbf{X}_1)&\ldots&\textbf{K}_{1,M}(\textbf{X}_1,\textbf{X}_M)\\
			\vdots&\ddots&\vdots\\
			\textbf{K}_{M,1}(\textbf{X}_M,\textbf{X}_1)&\ldots&\textbf{K}_{M,M}(\textbf{X}_M,\textbf{X}_M)
		\end{pmatrix}
		\label{mogp-covariance-matrix}
	\end{equation}
	has dimensions $(N_1+\cdots+N_M)\times(N_1+\cdots+N_M)$. This matrix is to be interpreted as having an $M\times M$ block structure
	with blocks of size $N_i\times N_j$ for any fidelity levels $i,j\in\{1,\ldots,M\}$. Each block is defined as:
	\begin{equation}
		\textbf{K}_{i,j}(\textbf{X}_i,\textbf{X}_j):=\begin{pmatrix}
			\kappa_{i,j}(\textbf{x}_{i,1},\textbf{x}_{j,1})&\ldots&\kappa_{i,j}(\textbf{x}_{i,1},\textbf{x}_{j,N_j})\\
			\vdots&\ddots&\vdots\\
			\kappa_{i,j}(\textbf{x}_{i,N_i},\textbf{x}_{j,1})&\ldots&\kappa_{i,j}(\textbf{x}_{i,N_i},\textbf{x}_{j,N_j})
		\end{pmatrix}.
		\label{mogp-block-matrix}
	\end{equation}
	Furthermore,
	\begin{equation}
		\boldsymbol{\mathcal{K}}(\textbf{x}, \boldsymbol{\mathcal{X}}):=\begin{pmatrix}
			\kappa_{1,1}(\textbf{x},\textbf{X}_1)&\ldots&\kappa_{1,M}(\textbf{x},\textbf{X}_1)\\
			\vdots&\ddots&\vdots\\
			\kappa_{M,1}(\textbf{x},\textbf{X}_M)&\ldots&\kappa_{M,M}(\textbf{x},\textbf{X}_M)
		\end{pmatrix}
	\end{equation}
	is an $(N_1+\cdots+N_M)\times M$ matrix consisting of stacked column vectors of length $N_m$, $m\in\{1,\ldots,M\}$, defined by
	\begin{equation}
		\kappa_{i,j}(\textbf{x},\textbf{X}_i):=\begin{pmatrix}
			\kappa_{i,j}(\textbf{x},\textbf{x}_{i,1})\\
			\vdots\\
			\kappa_{i,j}(\textbf{x},\textbf{x}_{i,N_i})
		\end{pmatrix}.
	\end{equation}
	
	A common structural assumption is the multiplicative separation between design and fidelity \cite{williams2007multi}: given a single-output covariance function $\kappa$ and a positive semi-definite matrix $\textbf{B}:=(b_{i,j})_{i,j=1\ldots,M}$, it can be postulated that $\boldsymbol{\mathcal{K}}\doteq\boldsymbol{\mathcal{K}}_{\text{MT}}(\kappa)=(\kappa_{i,j})_{i,j=1,\ldots,M}$ where
	\begin{equation}
		\kappa_{i,j}(\textbf{u},\textbf{v}):=b_{i,j}\cdot\kappa(\textbf{u},\textbf{v})
		\label{eq:mtask-covariance}
	\end{equation}
	for any designs $\textbf{u},\textbf{v}\in[0,1]^D$. The resulting type of MOGP is called a multi-task Gaussian process (MTGP). As a consequence, Equation \eqref{mogp-covariance-matrix} can be rewritten as a Hadamard product $\textbf{K}=\textbf{K}_{\text{Had}}\odot\textbf{B}$, where $\textbf{K}_{\text{Had}}:=(\textbf{K}(\textbf{X}_i,\textbf{X}_j))_{i,j=1\ldots,M}$ is again a block matrix with blocks defined in the sense of Equation \eqref{gp-prior}.
	
	Similar to single-fidelity Gaussian process regression, the goal of MOGP regression (MOGPR) is to find the maximum likelihood model parameter vector $\hat{\boldsymbol{\theta}}_{\text{MLE}}$ for a supposed covariance kernel. The expression for the maximum likelihood estimate of the model parameter vector is similar compared to the single-fidelity GP regression scenario in Equation \eqref{eq:mle}:
	\begin{equation}
		\hat{\boldsymbol{\theta}}_{\text{MLE}}=\argmin_{\boldsymbol{\theta}\in{\mathcal{T}}}\ln(\det(\textbf{K}_{\boldsymbol{\theta}}(\boldsymbol{\mathcal{X}})))+\textbf{y}^\top \textbf{K}^{-1}_{\boldsymbol{\theta}}(\boldsymbol{\mathcal{X}})\textbf{y}.
		\label{eq:mf-mle}
	\end{equation}	
	After locating $\hat{\boldsymbol{\theta}}_{\text{MLE}}$, the posterior predictive distribution of $\textbf{f}(\textbf{x})|(\textbf{Y}=\textbf{y})$ can be approximated by a regression-predictive, multivariate normal distribution with probability density function $\hat{\phi}$, in similar fashion to the single-fidelity scenario, cf. Equation \eqref{ppd-density-approx} and Equation \eqref{ppd-MLE}:
	\begin{equation}
		p(\textbf{f}(\textbf{x})|\textbf{Y}=\textbf{y})[y]\approx p(\textbf{f}(\textbf{x})|\textbf{Y}=\textbf{y},\boldsymbol{\Theta}=\hat{\boldsymbol{\theta}}_{\text{MLE}})[y]=\phi_{\hat{\boldsymbol{\mu}}(\textbf{x}),\hat{\boldsymbol{\Sigma}}(\textbf{x})}(y)=:\hat{\phi}(y;\textbf{x},\boldsymbol{\mathcal{D}},\boldsymbol{\mathcal{K}}),
	\end{equation}
	where
	$$\phi_{\boldsymbol{\mu},\boldsymbol{\Sigma}}(\textbf{z}):=\frac{1}{(2\pi)^{M/2}\sqrt{\det(\boldsymbol{\Sigma})}}\exp\left(-\frac12(\textbf{z}-\boldsymbol{\mu})^\top\boldsymbol{\Sigma}^{-1}(\textbf{z}-\boldsymbol{\mu})\right)$$
	denotes the multivariate normal probability density function with mean vector $\boldsymbol{\mu}$ and covariance matrix $\boldsymbol{\Sigma}$. Furthermore, $\hat{\boldsymbol{\mu}}:=\boldsymbol{\mu}_{\hat{\boldsymbol{\theta}}_{\text{MLE}}}$ (Equation \eqref{mogp-ppd-mean}) and $\hat{\boldsymbol{\Sigma}}:=\boldsymbol{\Sigma}_{\hat{\boldsymbol{\theta}}_{\text{MLE}}}$ (Equation \eqref{mogp-ppd-var}) are the mean vector and covariance matrix of the posterior distribution, parameterized by the maximum likelihood model parameter vector. Finally, it is useful to define
	\begin{equation}
		\hat{\phi}_m:=\phi_{\hat{\mu}_m(\textbf{x}),\hat{\sigma}^2_m(\textbf{x})}.
	\end{equation}
	
	\subsection{Multi-fidelity Bayesian optimization with MOGPR}
	\label{sec:mfbo}
	
	The goal of (single-fidelity) Bayesian optimization is to optimize an objective function $f$. When there are $M$ objectives $f_1,\ldots,f_M$, ordered in increasing levels of fidelity, the commonly assumed goal is to optimize only $f_M$ \cite{kandasamy2017multi, zhang2018variable, jiang2019variable, do2023multi}. In particular, the interest does \textit{not} lie with optimizing $f_1,\ldots,f_{M-1}$, and data acquired from any of these $M-1$ objectives merely serve an auxiliary purpose to transfer knowledge to fidelity level $M$. This draws an important but rather confusing distinction between multi-output BO (MOBO) and BO that utilizes MOGP regression models, which bears the name multi-fidelity Bayesian optimization (MFBO) in this manuscript: while MOBO tries to optimize the various objective functions simultaneously \cite{maddox2021bayesian, swersky2013multi, laumanns2002bayesian}, MFBO leverages the knowledge transfer facilitated by MOGPR to optimize $f_M$, and is the focal point of this manuscript.
	
	In the context of multi-fidelty Bayesian optimization, it is important to address the literature surrounding acquisition functions capable of handling multi-fidelity data sets. A large class of acquisition functions involve the arithmetic combination of functions that pertain to various factors involving the design space and/or the fidelity space. In particular, ideas from single-fidelity acquisition functions get exploited and augmented to handle multiple levels of fidelity. This is similar to how constrained EI \cite{schonlau1997computer} or local penalization of generic acquisitions \cite{gonzalez2016batch} is constructed in the single-fidelity scenario. 
	
	An early example of a multiplicative acquisition is an augmented version of the EI acquisition function based on multi-fidelity sequential kriging optimization \cite{huang2006sequential}, also called variable-fidelity EI (VF-EI). It is possible to modify VF-EI by using the logarithmic EI acquisition function instead \cite{ament2023unexpected}. This multi-fidelity acquisition function multiplies the EI function applied to the highest fidelity, with fidelity-dependent parameters such as the ratio of computational cost and correlation between the highest fidelity and the fidelity in question. Other similar multi-fidelity acquisition functions built on top of single-fidelity improvement-based acquisitions include extended expected improvement \cite{liu2018sequential}, variable-fidelity probability of improvement (VF-PI) \cite{ruan2020variable} and variable-fidelity upper confidence bound (VF-UCB) \cite{jiang2019variable}. For this manuscript, the main focus will be on the VF-(Log)EI and VF-UCB. See Equation \eqref{eq:vfei} and Equation \eqref{eq:vfucb}.
	
	\begin{align}
		\alpha_{\text{VF-EI}}(\textbf{x},m;\hat{\phi}^{(i)})&:=\alpha_{\text{EI}}(\textbf{x};\hat{\phi}^{(i)}_m)\cdot\text{CR}(m)\cdot\rho(m);
		\label{eq:vfei}\\
		\alpha_{\text{VF-UCB}}(\textbf{x},m;\hat{\phi}^{(i)})&:=\omega_1\cdot\hat{\mu}^{(i)}_m(\textbf{x})+\omega_2\cdot\hat{\sigma}^{(i)}_m(\textbf{x})\cdot\text{CR}(m).
		\label{eq:vfucb}
	\end{align}
	Here, $\text{CR}(m)$ stands for the cost ratio, i.e. the ratio of (computational) expense of one evaluation of $f_m$ compared to one evaluation of $f_M$. Furthermore, in the case of $M=2$, $\rho(m=1)$ is the Pearson correlation coefficient between a the low- and high-fidelity objectives, otherwise $\rho(m=2)=1$. Finally, in the case of VF-UCB, $\omega_1,\omega_2$ are weights that depend on the coefficients of variation pertaining to evaluations of $f_m$ throughout the entire design space.
	
	The Bayesian optimization algorithm in the multi-fidelity scenario is largely similar to Algorithm \ref{algo:bo}, and is presented in Algorithm \ref{algo:mfbo}.
	
	\begin{algorithm}[H]
		\caption{Bayesian optimization with MFGPR (MFBO)}
		\begin{algorithmic}[1]
			\Require{Design of train experiments $\boldsymbol{\mathcal{D}}^{(0)}$, covariance function $\boldsymbol{\mathcal{K}}$, model parameter vector domain $\mathcal{T}$, acquisition function $\alpha$, number of iterations $I$, cost function $c$, budget $B$}
			\State $b\gets0$
			\For{$i=1,\ldots,I$}
			\If{$b>B$}
			\State \textbf{break}
			\EndIf
			\State\small$\hat{\boldsymbol{\theta}}_{\text{MLE}}^{(i)}\gets\argmin_{\boldsymbol{\theta}\in\mathcal{T}}\ln(\det(\textbf{K}_{\boldsymbol{\theta}}(\boldsymbol{\mathcal{X}}^{(i-1)})))+\textbf{y}^{(i-1)\top} \textbf{K}^{-1}_{\boldsymbol{\theta}}(\boldsymbol{\mathcal{X}}^{(i-1)}) \textbf{y}^{(i-1)}$
			\State $\hat{\phi}^{(i)}\gets\hat{\boldsymbol{\theta}}_{\text{MLE}}^{(i)}$
			\State $(\textbf{x}^{(i)\top},m^{(i)})\gets\argmax_{\textbf{x}\in[0,1]^D,m\in\{1,\ldots,M\}}\alpha(\textbf{x},m;\hat{\phi}^{(i)})$
			\State $y^{(i)}\gets f_{m^{(i)}}(\textbf{x}^{(i)})$
			\State $b\gets b+c(m^{(i)})$
			\If{$m=m^{(i)}$}
			\State $\textbf{D}_m^{(i)}\gets(\textbf{D}_m^{(i-1)},(\textbf{x}^{(i)\top},y^{(i)}))^\top$
			\Else
			\State $\textbf{D}_m^{(i)}\gets\textbf{D}_m^{(i-1)}$
			\EndIf
			\EndFor
			\State $(\textbf{x}_{\text{rec}},y_{\text{rec}})\gets\text{Rec}(\boldsymbol{\mathcal{D}}^{(I)})$
			\\\Return $(\textbf{x}_{\text{rec}},y_{\text{rec}})$
		\end{algorithmic}
		\label{algo:mfbo}
	\end{algorithm}
	
	One main difference between single- and multi-fidelity BO is the augmentation of the data set at each fidelity, showcased by steps 11-15. If fidelity $m$ is selected to represent the fidelity level to sample at for iteration $i$, only $\textbf{D}^{(i)}_m$ is updated, while all the data sets at the other fidelities remain unchanged. Similar to single-fidelity BO, the multi-fidelity counterpart has a recommendation function which depends on the final multi-fidelity data set after the final iteration $I$. This recommender returns the largest value among $\{f_M(\textbf{x}_1^*),f_M(\textbf{x}_2^*),\ldots,f_M(\textbf{x}_M^*)\}$ and the corresponding design parameter vector, where $\textbf{x}_m^*$ is the incumbent maximizer of $f_m$ after iteration $I$.
	
\begin{figure}[H]
		\centering
		\includesvg[width=\linewidth]{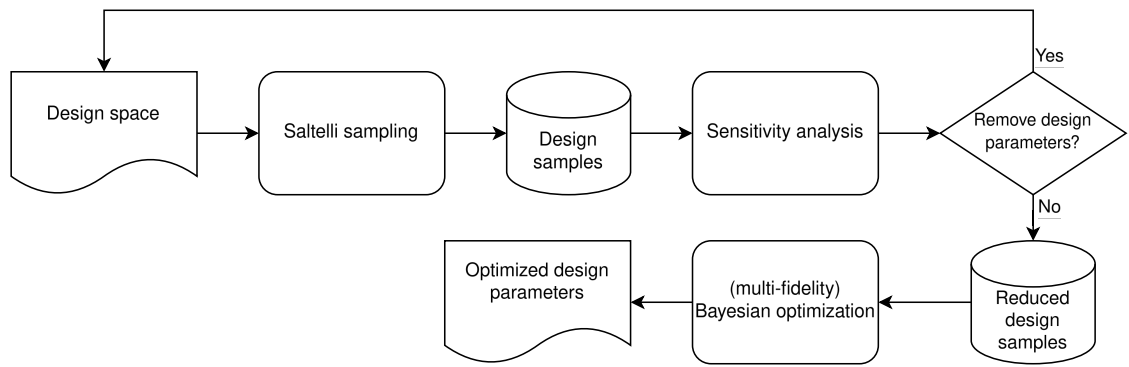}
		\caption{Flowchart diagram of the Bayesian data driven framework which includes BO (Figure \ref{fig:bo-diagram}).}
		\label{fig:bayesian_framework}
	\end{figure}

	As a summary, the concepts of Sobol' sensitivity analysis (Section \ref{sec:ssa}) and Bayesian optimization (this section) are combined into one Bayesian data-driven framework. See Figure \ref{fig:bayesian_framework} for a schematic overview of the framework.
	
	\section{Problem description and data analysis}
	\label{spinodal-description}
	
	In order to optimize the energy absorption objective of the spinodoid structure according to the data-driven framework in Figure \ref{fig:bayesian_framework}, a number of relevant design parameters have been identified, along with their domains. See Table \ref{tab:4D-EA-design} for an overview.
	
	\begin{table}[H]
		\centering
		\caption{Design parameters for the 4D EA optimization problem.}
		\begin{tabular}{|lll|}
			\hline
			Design parameter & Lower bound & Upper bound \\ \hline
			$\rho$              & 0.3        & 0.6       \\ \hline
			$\theta_1$              & 0°    & 90°       \\ \hline
			$\theta_2$             & 0°    & 90°    \\ \hline
			$\theta_3$             & 0°    & 90°   \\ \hline
		\end{tabular}
		\label{tab:4D-EA-design}
	\end{table}
	
\begin{figure}[H]
		\centering
		\begin{subfigure}{\textwidth}
			\includegraphics[width=\textwidth]{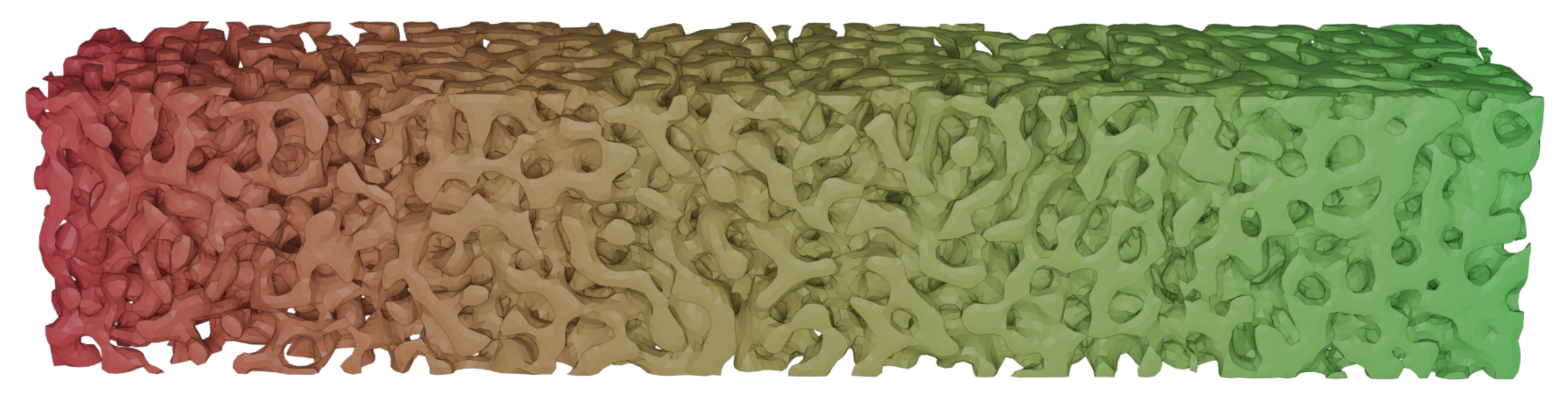}
		\end{subfigure}
		\caption{Varying levels of $\rho\in[0.3, 0.6]$, increasing from left to right, for fixed values $\theta_1=90^\circ$, $\theta_2=0^\circ$ and $\theta_3=0^\circ$.}
		\label{fig:spinodal-rho}
	\end{figure}

	From a physical point of view, $\rho$ represents the relative density of the structure as a material to air ratio and the angles $\theta_1,\theta_2,\theta_3$ give rise to the configuration of the structure itself \cite{kansara2024multi,cahn1965phase,zheng2021data}. See Figure \ref{fig:spinodal-rho} for an illustration\footnote{It should be noted that the depicted structure serves as a visual aid and the values $\rho$ is kept constant throughout the structure for any and all data-related results.}.
	
	\subsection{Objective}
	As the one-dimensional compressive loading simulation is performed, a force $P$ is applied across a displacement $x$ until a threshold $\delta_{\max}$ is achieved. As the material deforms, the compression force needed might increase or decrease. From this, a force-displacement behavior $P(x)$ can be observed. 
	
	The relevant data-driven objective related to the scope of this problem is the normalized energy absorption value, which equals the integral of the force across the continuous displacement variable, divided by the EA value obtained from a cube made out of solid material with relative density $\rho$, denoted as $\text{EA}_s$.
	\begin{equation}
		\text{EA}=\frac{1}{\text{EA}_s}\int_0^{\delta_{\max}}P(x)\,\mathrm{d}x.
		\label{eq:EA}
	\end{equation}
	
	\subsection{4D data}
	
	The starting point of this data-driven analysis is the sampling and visualization of data. The sampling is performed such that it is compatible with performing SSA. 
	
\begin{figure}[H]
		\centering
		\begin{subfigure}{.45\textwidth}
			\includegraphics[width=\textwidth]{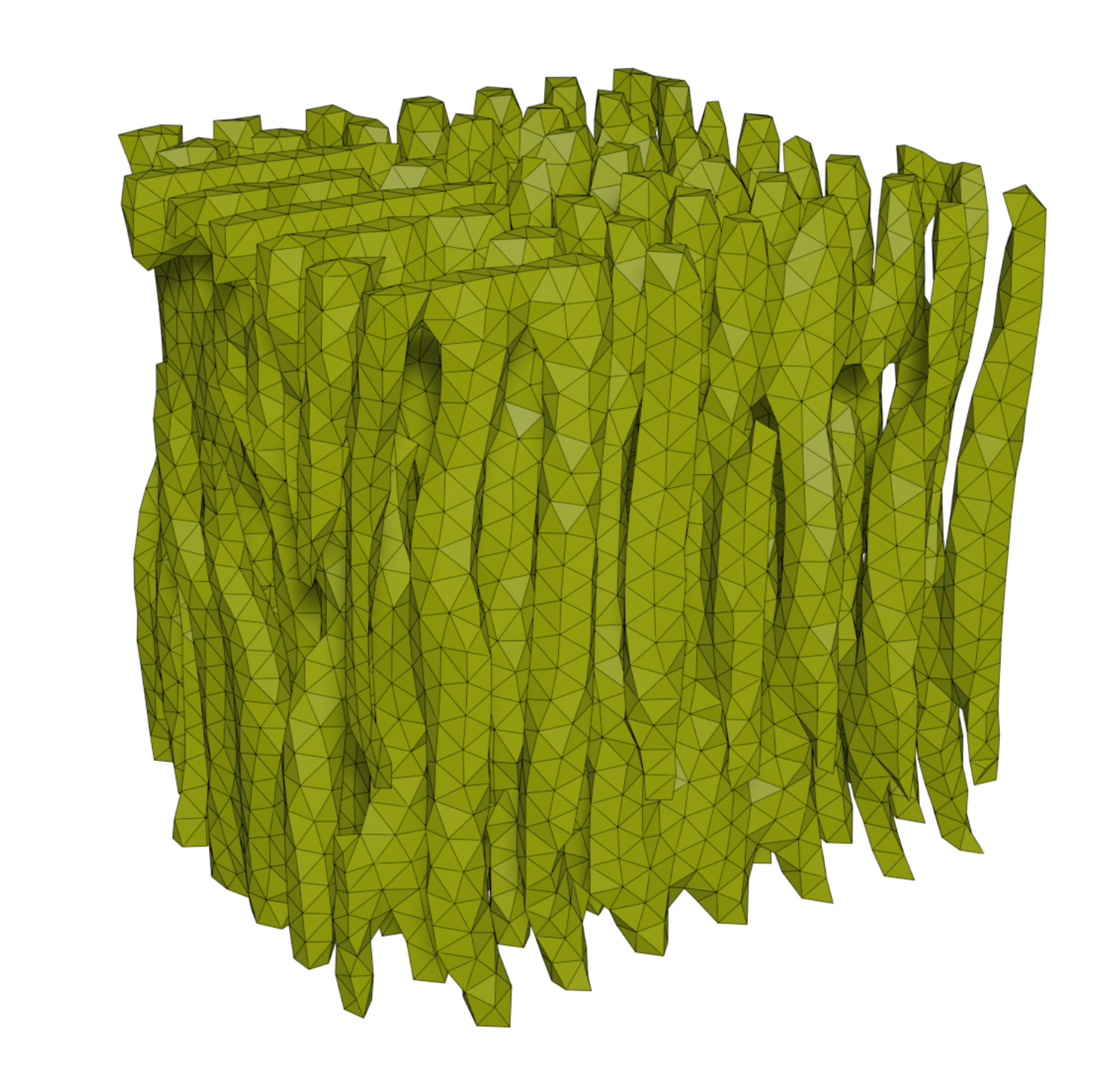}
			\caption{Mesh resolution 20}
			\label{fig:spinodal-meshes-lf}
		\end{subfigure}
		\begin{subfigure}{.45\textwidth}
			\includegraphics[width=\textwidth]{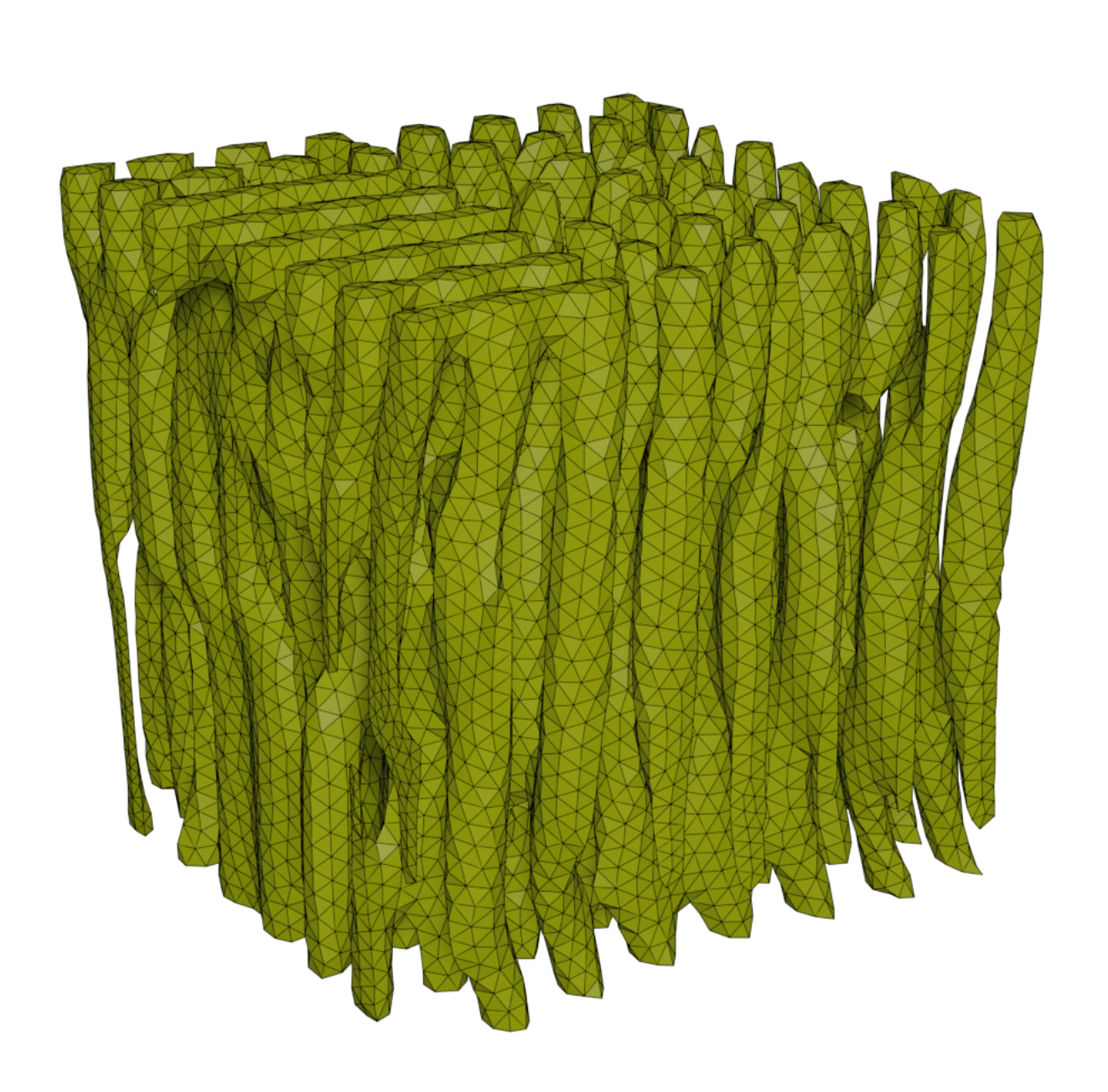}
			\caption{Mesh resolution 30}
			\label{fig:spinodal-meshes-hf}
		\end{subfigure}
		\caption{Two mesh resolutions for fixed values $\rho=0.3$, $\theta_1=15^\circ$, $\theta_2=15^\circ$ and $\theta_3=0^\circ$. The coarse mesh represents the low-fidelity model, while the fine mesh is considered high-fidelity. Intuitively, a low value for $\theta_3$ arranges the structure into columns aligned into the loading direction.}
		\label{fig:spinodal-meshes}
	\end{figure}

	Naturally, the computational cost depends on the mesh resolution at which the spinodoid structure is generated; it is expected that the fine mesh (Figure \ref{fig:spinodal-meshes-hf}) simulations will take longer than the coarse mesh (Figure \ref{fig:spinodal-meshes-lf}) simulations. 
	
\begin{figure}[H]
		\centering
		\includesvg[width=.8\textwidth]{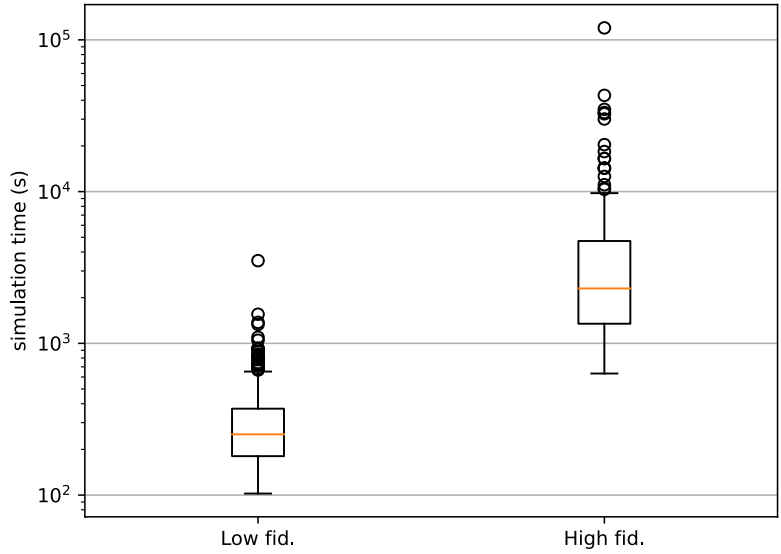}
		\caption{Simulation times for mesh resolution 20 (left) and 30 (right). The median simulation time when the coarser mesh is applied is 251 seconds, which is a factor 0.11 of the median simulation time of the high-fidelity samples (2300 seconds).}
		\label{fig:spinodal-simulation-times}
	\end{figure}

	It should be noted that the simulation cost depends on the design parameters. In order to empirically estimate how much the computational cost differs between the objective at the two fidelity levels, sets of Saltelli samples are used to generate a statistic of computational run times, presented in Figure \ref{fig:spinodal-simulation-times}. For the high-fidelity objective, 96 design samples are used, whereas the low-fidelity objective is evaluated on the same design samples, and an additional 288 samples, for a total of 386 samples.
	
	Given the long simulation times compared to the duration of a BO iteration with the same computational power, it makes sense to consider Bayesian data-driven techniques to optimize the spinodal structural design for its energy absorption.
	
\begin{figure}[H]
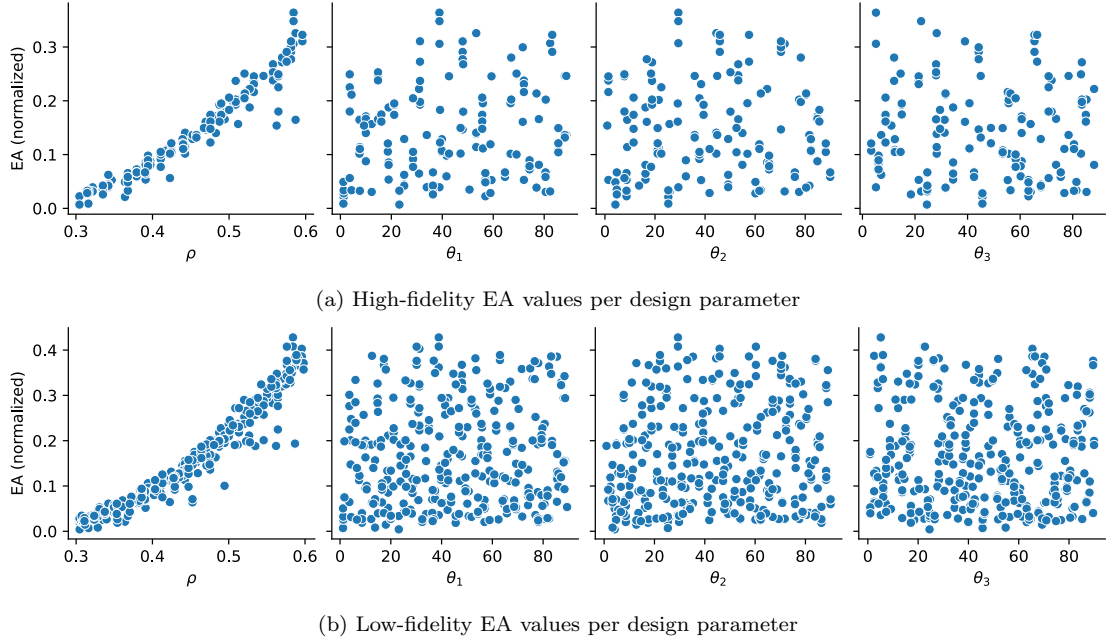

		\centering
		\begin{subfigure}{\textwidth}
			\includesvg[width=\textwidth]{img/applications/4D_data_plot_hf.svg}
			\caption{High-fidelity EA values per design parameter}
			\label{fig:spinodal-EA-4D-data_plot-hf}
		\end{subfigure}
		\begin{subfigure}{\textwidth}
			\includesvg[width=\textwidth]{img/applications/4D_data_plot_lf.svg}
			\caption{Low-fidelity EA values per design parameter}
			\label{fig:spinodal-EA-4D-data_plot-lf}
		\end{subfigure}
		\caption{The collection of Saltelli samples used in Figure \ref{fig:spinodal-simulation-times} and their normalized EA value projections onto the design parameter axes.}
		\label{fig:spinodal-EA-4D-data_plot}
	\end{figure}

	As a next step in the data analysis, the aim is to analyze the influence that each of the design parameters has with respect to the objective. See Figure \ref{fig:spinodal-EA-4D-data_plot}.
	
	The leftmost scatter plot in both Figure \ref{fig:spinodal-EA-4D-data_plot-hf} and Figure \ref{fig:spinodal-EA-4D-data_plot-lf}  appears to show a strong relationship between $\rho$ and the normalized EA value, whereas the projections onto the $\theta$-dimensions exhibit much less structure. This holds for both the low- and high-fidelity objectives.
	
\begin{figure}[H]
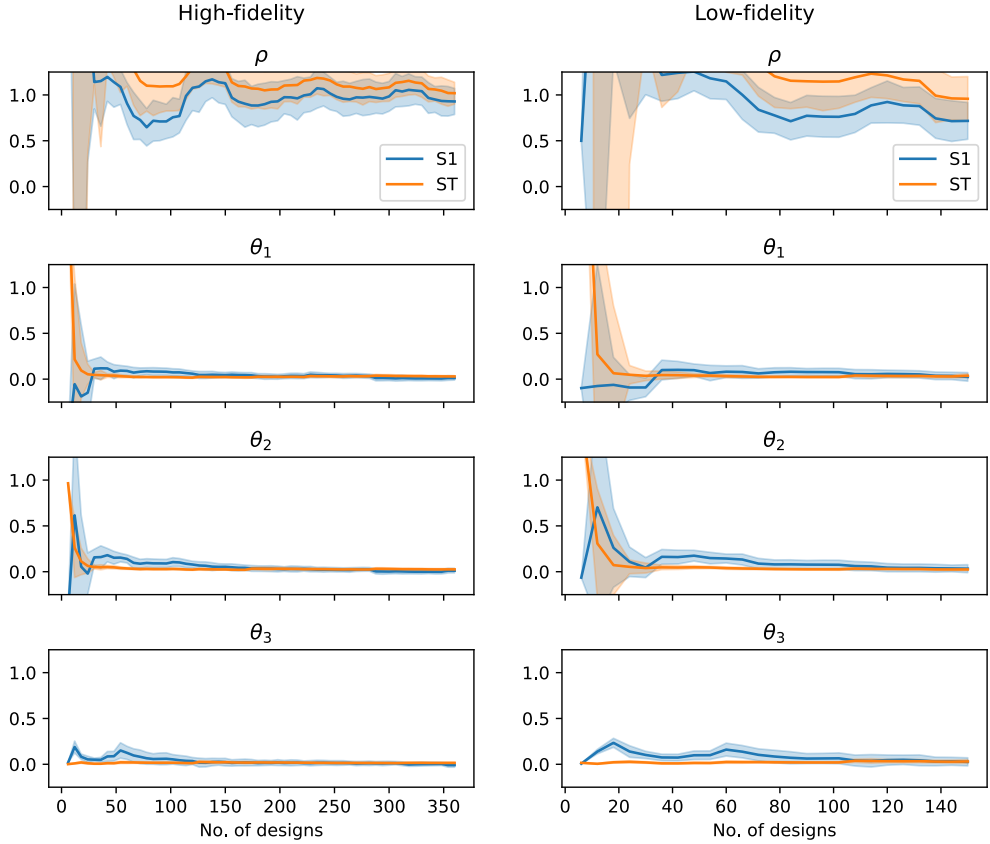

		\centering
		\begin{subfigure}{.45\textwidth}
			\includesvg[width=\textwidth]{img/applications/EA-4D-hf-conv.svg}
		\end{subfigure}
		\begin{subfigure}{.45\textwidth}
			\includesvg[width=\textwidth]{img/applications/EA-4D-lf-conv.svg}
		\end{subfigure}
		\caption{First (S1) and total (ST) order Sobol' sensitivity indices and corresponding bootstrapped confidence interval per design parameter as a function of the number of Saltelli designs.}
		\label{fig:spinodal-EA-4D-sensitivity-indices-convergence}
	\end{figure}

	In order to confirm this visual heuristic, SSA can be performed on the readily available Saltelli samples. The first and total order total sensitivity indices are calculated with different quantities of design parameter vectors, in order to observe any stabilization behavior exhibited by the sensitivity indices. See Figure \ref{fig:spinodal-EA-4D-sensitivity-indices-convergence}.
	
	It can be concluded from Figure \ref{fig:spinodal-EA-4D-sensitivity-indices-convergence} that the sensitivity indices are stabilizing when more than 150 design samples are utilized. The sensitivity of both low- and high-fidelity EA objectives with respect to $\rho$ approaching 1, while the $S_i$ and $S_{T_i}$ values of $\theta_1$, $\theta_2$, $\theta_3$ approach 0. This validates the initial observations made from Figure \ref{fig:spinodal-EA-4D-data_plot}.
	
	Given the findings related to the 4D design space, the amount of input-sensitivity and input-output structure revealed with regards to the angular design parameters is limited. Therefore, the next step is to select a fixed value for $\rho$ in order to observe the residual variance-based sensitivity among $\theta_1,\theta_2,\theta_3$. 
	
	\subsection{3D data}
\begin{figure}[H]
		\centering
		\begin{subfigure}{\textwidth}
			\includesvg[width=\textwidth]{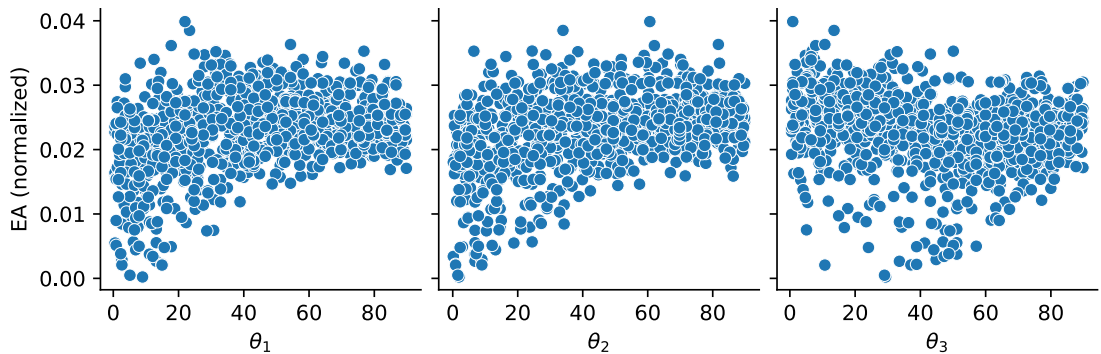}
			\caption{High-fidelity EA values per design parameter}
			\label{fig:spinodal-EA-3D-data_plot-hf}
		\end{subfigure}
		\begin{subfigure}{\textwidth}
			\includesvg[width=\textwidth]{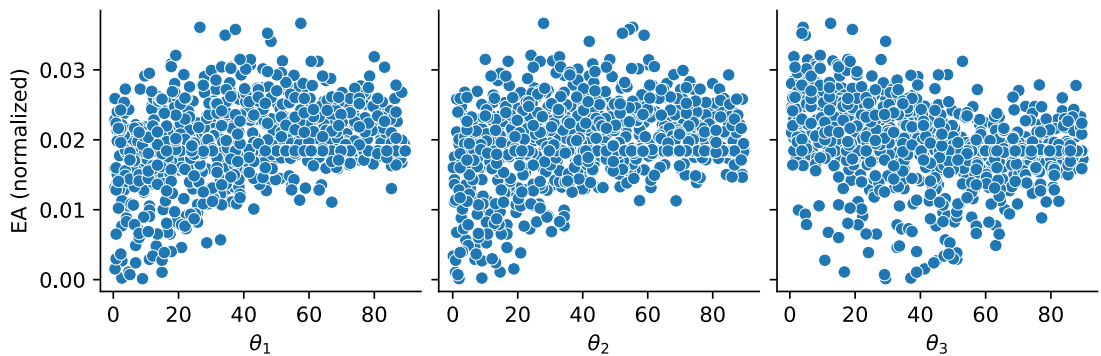}
			\caption{Low-fidelity EA values per design parameter}
			\label{fig:spinodal-EA-3D-data_plot-lf}
		\end{subfigure}
		\caption{The collection of 980 Saltelli samples (each fidelity) and their normalized EA value projections onto the remaining design parameter axes when $\rho=0.3$ is fixed.}
		\label{fig:spinodal-EA-3D-data_plot}
	\end{figure}

	Upon viewing the axial objective projections in Figure \ref{fig:spinodal-EA-4D-data_plot} and the sensitivity profiles in Figure \ref{fig:spinodal-EA-4D-sensitivity-indices-convergence}, it cannot be ruled out that one of the $\theta$-parameters could be more sensitive than the others. The investigation therefore continues by performing a variance-based sensitivity analysis over the remaining design parameters $\theta_1,\theta_2,\theta_3$, with a fixed value for $\rho=0.3$. This lower bound value for $\rho$ is chosen for mass scaling reasons: a reduction in the FE simulation times expedites the process of design space sampling. Similar to the 4D case, a coordinate-wise projection scatter plot was produced. See Figure \ref{fig:spinodal-EA-3D-data_plot}.
	
\begin{figure}[H]
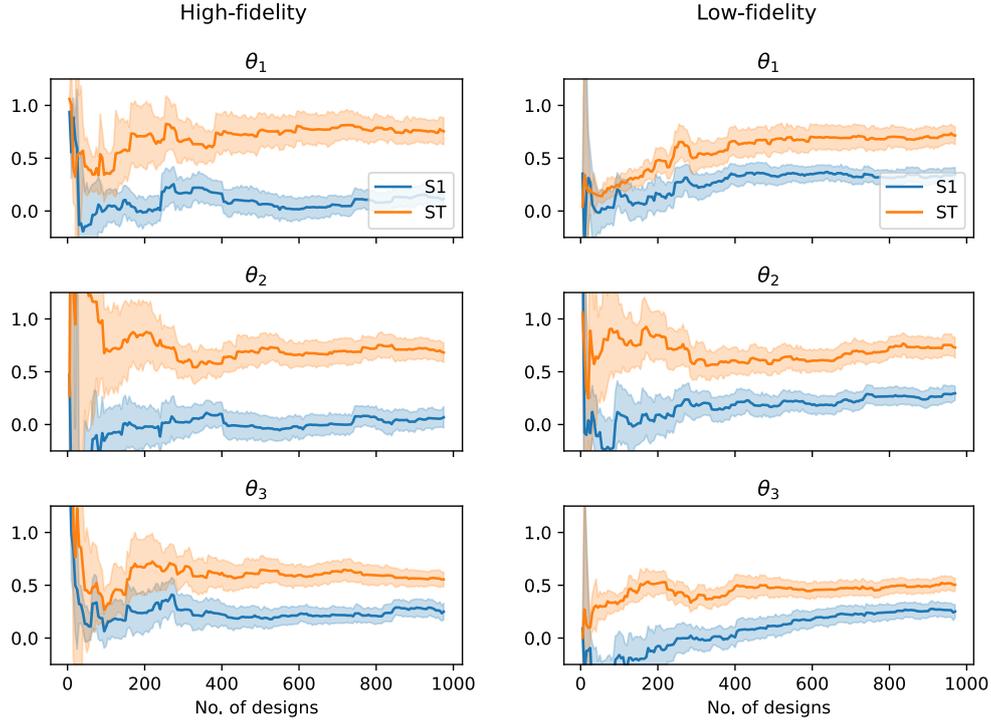

		\centering
		\begin{subfigure}{.45\textwidth}
			\includesvg[width=\textwidth]{img/applications/EA-3D-hf-conv.svg}
		\end{subfigure}
		\begin{subfigure}{.45\textwidth}
			\includesvg[width=\textwidth]{img/applications/EA-3D-lf-conv.svg}
		\end{subfigure}
		\caption{First (S1) and total (ST) order Sobol' sensitivity indices of the 3D EA objective (fixed $\rho=0.3$), and corresponding bootstrapped confidence interval per design parameter as a function of the number of Saltelli designs. The discrepancy between the first and total-order indices for all design parameters indicates that higher-order joint effects are statistically significant with respect to varying the objective.}
		\label{fig:spinodal-EA-3D-sensitivity-indices-convergence}
	\end{figure}

	Finally, the same sensitivity index convergence analysis is performed that gave rise to Figure \ref{fig:spinodal-EA-4D-sensitivity-indices-convergence} in the 4D scenario. The resulting observations are shown in Figure \ref{fig:spinodal-EA-3D-sensitivity-indices-convergence}.
	
	From Figure \ref{fig:spinodal-EA-3D-sensitivity-indices-convergence}, it can be seen that the angular design parameters that give rise to the spinodoid structure are all approximately equally sensitive towards energy absorption. Moreover, it is noted that the sensitivity indices take longer to stabilize. With both mesh resolutions, the indices only stabilize when more than 400 design samples are used, a significant increase compared to the 150 needed for the case with the 4D design space. This indicates the difficulty for the SSA procedure to determine the sensitivity indices due to less structure in the overall data, as confirmed visually in Figure \ref{fig:spinodal-EA-3D-data_plot}.
	
\begin{figure}[H]
		\centering
		\begin{subfigure}{.75\textwidth}
			\includesvg[width=\textwidth]{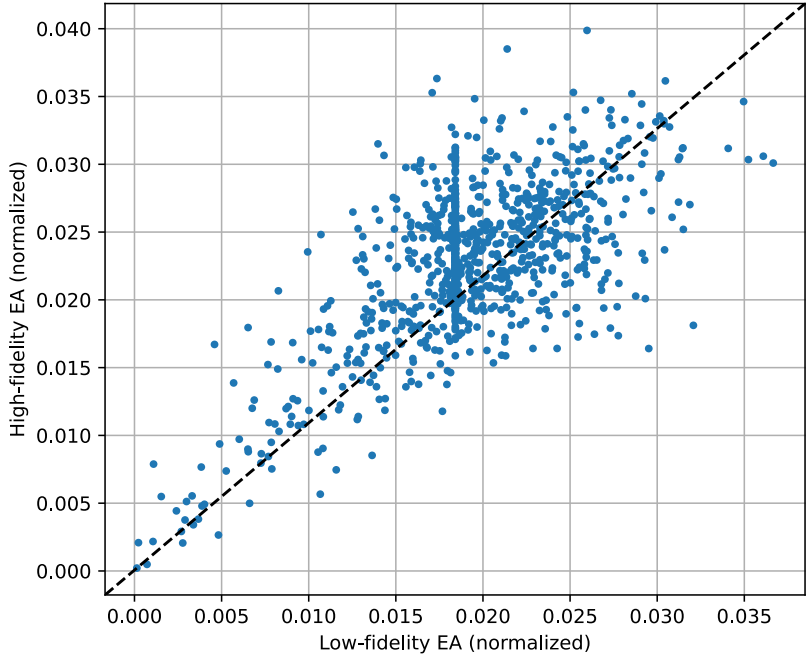}
		\end{subfigure}
		\caption{Low- and high-fidelity EA values at the coincident 3D Saltelli design samples used in Figure \ref{fig:spinodal-EA-3D-sensitivity-indices-convergence}. The correlation coefficient between the two sets of data points is approximately 0.68. It should be noted that the cluster of designs with low-fidelity EA values at around 0.018 is due to a numerical artifact resulting from the FE meshing procedure, resulting in the same EA value for a range of different $\theta$-values.}
		\label{fig:spinodal-EA-3D-interfidelity-EA}
	\end{figure}

	Furthermore, it is of interest to investigate the level of correlation between the low- and high-fidelity data points. See Figure \ref{fig:spinodal-EA-3D-interfidelity-EA}.
	
	As is visible from Figure \ref{fig:spinodal-EA-3D-interfidelity-EA}, a moderate level of (linear) correlation exists between the low- and high-fidelity objectives. This motivates the application of multi-fidelity Bayesian techniques, due to the transfer learning potential from the low- to high-fidelity levels.
	The high computational cost in addition to this observed non-triviality of the EA-objective landscape with respect to the angular design parameter makes the reduced three-dimensional problem interesting for Bayesian optimization.
	
	\section{Optimization results}
	\subsection{Single-fidelity BO}
	
\begin{figure}[H]
		\centering
		\includesvg[width=\textwidth]{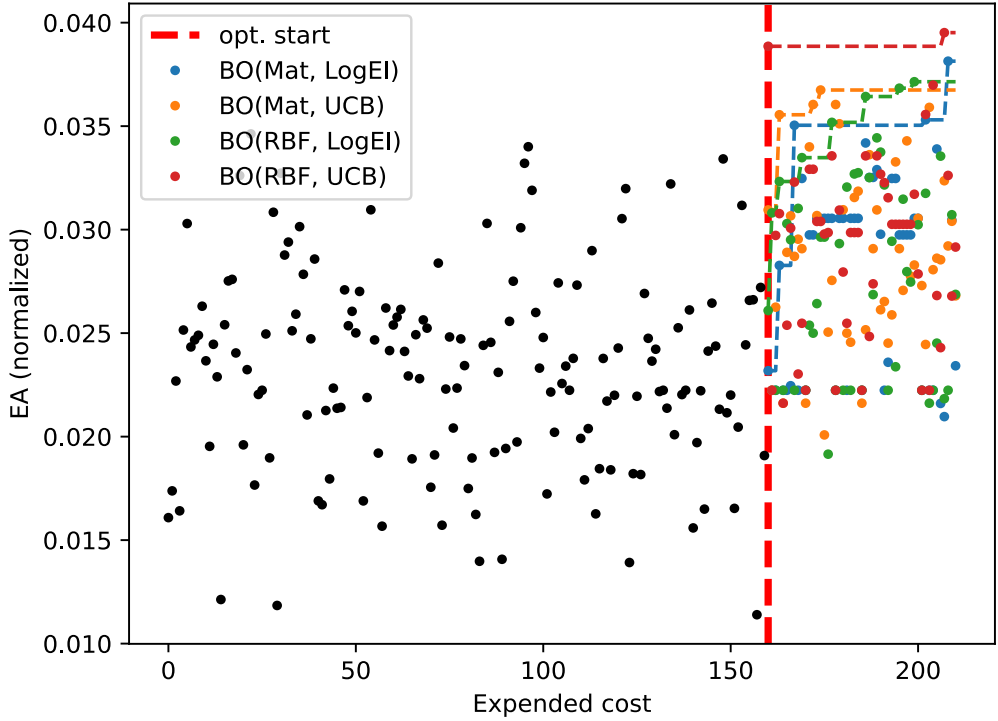}
		\caption{A number of standard BO histories (colored dots) and each cumulative maximum (colored dashed line). The black dots correspond to the common initial DoE, whereas the vertical thick red dashed line indicates the cost value at which BO was initiated. The optimization was allowed to run for a computational cost equivalent of 50 high-fidelity evaluations.}
		\label{fig:spinodal-EA-3D-opt}
	\end{figure}

	Bayesian optimization (Algorithm \ref{algo:bo}) with four combinations of covariance functions (Matérn kernel, RBF kernel) and acquisition functions (Logarithmic EI, UCB) are applied to maximize the normalized EA. For this experiment, the computational budget for the initial DoE is equivalent to 160 high-fidelity evaluations, while the optimization budget is 50 high-fidelity equivalents. The optimization trajectories are shown in Figure \ref{fig:spinodal-EA-3D-opt}.
	
	\begin{table}[H]
		\centering
		\caption{Optima to the 3D EA optimization problem found by various single-fidelity BO configurations. The best design is indicated in boldface.}
		\begin{tabular}{|llllll|}
			\hline
			\multirow{2}{*}{Acquisition}             & \multirow{2}{*}{Kernel} & \multicolumn{4}{c|}{Optimized values}                                                                                 \\ \cline{3-6} 
			&                         & \multicolumn{1}{l}{$\theta_1$} & \multicolumn{1}{l}{$\theta_2$} & \multicolumn{1}{l}{$\theta_3$} & EA (normalized) \\ \hline
			\multirow{2}{*}{$\alpha_{\text{LogEI}}$} & $\kappa_{\text{Mat}}$   & \multicolumn{1}{l}{$46.6^\circ$}      & \multicolumn{1}{l}{$56.5^\circ$}      & \multicolumn{1}{l}{$10.0^\circ$}      & 0.038139        \\ \cline{2-6} 
			& $\kappa_{\text{RBF}}$   & \multicolumn{1}{l}{$41.7^\circ$}      & \multicolumn{1}{l}{$56.7^\circ$}      & \multicolumn{1}{l}{$0.0^\circ$}       & 0.037146        \\ \hline
			\multirow{2}{*}{$\alpha_{\text{UCB}}$}   & $\kappa_{\text{Mat}}$   & \multicolumn{1}{l}{$36.8^\circ$}      & \multicolumn{1}{l}{$62.9^\circ$}      & \multicolumn{1}{l}{$0.0^\circ$}       & 0.037123        \\ \cline{2-6} 
			& $\kappa_{\text{RBF}}$   & \multicolumn{1}{l}{$\mathbf{37.6}^\circ$}      & \multicolumn{1}{l}{$\mathbf{61.7}^\circ$}      & \multicolumn{1}{l}{$\mathbf{0.0}^\circ$}       & \textbf{0.039517}        \\ \hline
		\end{tabular}
		\label{tab:spinodal-EA-3D-opt-rec}
	\end{table}
	
	From Table \ref{tab:spinodal-EA-3D-opt-rec}, it can be seen that all optimizers prefer designs with a low values of $\theta_3$. This indicates a higher value of energy absorption when the spinodoid structure assumes a columnar shape, such as the ones displayed in Figure \ref{fig:spinodal-meshes}. 
	
	\subsection{Multi-fidelity BO}
	As explained in Section \ref{spinodal-description}, it is possible to exploit the cost-accuracy trade-off that naturally arises with mesh resolution when FE simulations are concerned. With a small low- to high-fidelity cost ratio (0.11), it is therefore sensible to use multi-fidelity Bayesian optimization (Algorithm \ref{algo:mfbo}) for solving the energy absorption maximization problem. 
	
\begin{figure}[H]
		\centering
		\includesvg[width=\textwidth]{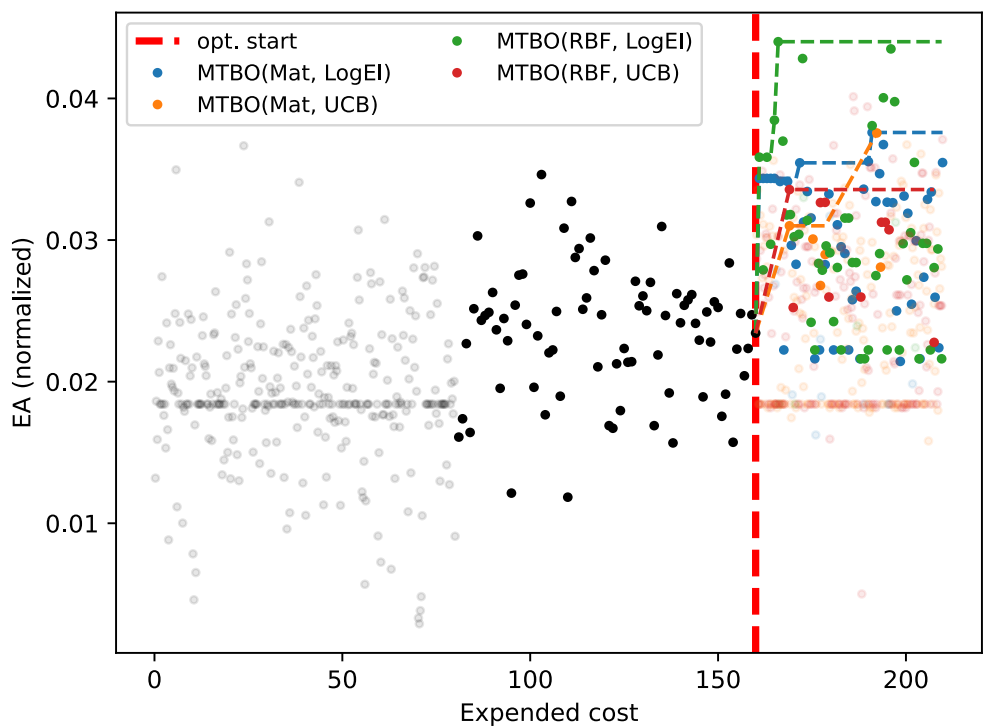}
		\caption{A number of EA optimization histories with multi-task Bayesian optimization. The solid dots indicate high-fidelity data points, while the faded dots correspond to low-fidelity data. The evaluated EA value is plotted across the expended computational cost in units of high-fidelity data points. The black dots represent the initial high-fidelity data, shared across all BO runs.}
		\label{fig:spinodal-EA-3D-mf-opt}
	\end{figure}

	For this numerical experiment, a total computational cost of 160 high-fidelity evaluations is spent on the initial data set. The same optimization budget of 50 high-fidelity evaluation equivalents is used, compared to the single-fidelity BO experiment in Figure \ref{fig:spinodal-EA-3D-opt}. The multi-task Bayesian optimization (MTBO) objective histories are shown in Figure \ref{fig:spinodal-EA-3D-mf-opt}. 
	
	\begin{table}[H]
		\centering
		\caption{Optima to the 3D EA optimization problem found by various MTBO configurations. The best design is indicated in boldface.}
		\begin{tabular}{|llllll|}
			\hline
			\multirow{2}{*}{Acquisition}             & \multirow{2}{*}{Kernel} & \multicolumn{4}{c|}{Optimized values}                                                                                 \\ \cline{3-6} 
			&                         & \multicolumn{1}{l}{$\theta_1$} & \multicolumn{1}{l}{$\theta_2$} & \multicolumn{1}{l}{$\theta_3$} & EA (normalized) \\ \hline
			\multirow{2}{*}{$\alpha_{\text{LogEI}}$} & $\kappa_{\text{Mat}}$   & \multicolumn{1}{l}{$44.1^\circ$}      & \multicolumn{1}{l}{$46.2^\circ$}      & \multicolumn{1}{l}{$8.9^\circ$}       & 0.040766        \\ \cline{2-6} 
			& $\kappa_{\text{RBF}}$   & \multicolumn{1}{l}{$\mathbf{44.4}^\circ$}      & \multicolumn{1}{l}{$\mathbf{43.9}^\circ$}      & \multicolumn{1}{l}{$\mathbf{2.4}^\circ$}       & \textbf{0.044007}        \\ \hline
			\multirow{2}{*}{$\alpha_{\text{UCB}}$}   & $\kappa_{\text{Mat}}$   & \multicolumn{1}{l}{$41.6^\circ$}      & \multicolumn{1}{l}{$50.2^\circ$}      & \multicolumn{1}{l}{$3.9^\circ$}       & 0.040872        \\ \cline{2-6} 
			& $\kappa_{\text{RBF}}$   & \multicolumn{1}{l}{$44.2^\circ$}      & \multicolumn{1}{l}{$45.2^\circ$}      & \multicolumn{1}{l}{$6.0^\circ$}       & 0.038878        \\ \hline
		\end{tabular}
		\label{tab:spinodal-EA-3D-mf-opt-rec}
	\end{table}
	
	From Table \ref{tab:spinodal-EA-3D-mf-opt-rec}, it can be seen that the values that multi-task BO recommend are generally better than the single-fidelity BO counterparts, presented in Table \ref{tab:spinodal-EA-3D-opt-rec}. Importantly, it can be seen that there is an 11\% overall increase in the optimally located normalized EA value across all kernels and acquisitions, while there is an average 8.6\% increase among methods with the same covariance and acquisition function types. This latter fact can be attributed to the knowledge transfer that occurs from low- to high-fidelity. This transfer is captured by the multi-task covariance kernel and the multi-fidelity acquisition functions, supported by the substantial correlation between the low- and high-fidelity data, as noted from Figure \ref{fig:spinodal-EA-3D-interfidelity-EA}.
	
	Moreover, it should be noted that three out of four values in Table \ref{tab:spinodal-EA-3D-opt-rec} are strongly columnar, with optimal values of $\theta_3=0^\circ$. This stands in contrast to the slightly higher values of $\theta_3$ found by all of the multi-task Bayesian optimizers ($2^\circ\sim9^\circ$). These structures  are bending-dominated structures, which are known to distribute stress across the cellular structure evenly \cite{kansara2024multi}.
	 
\begin{figure}[H]
		\centering
		\includesvg[width=\textwidth]{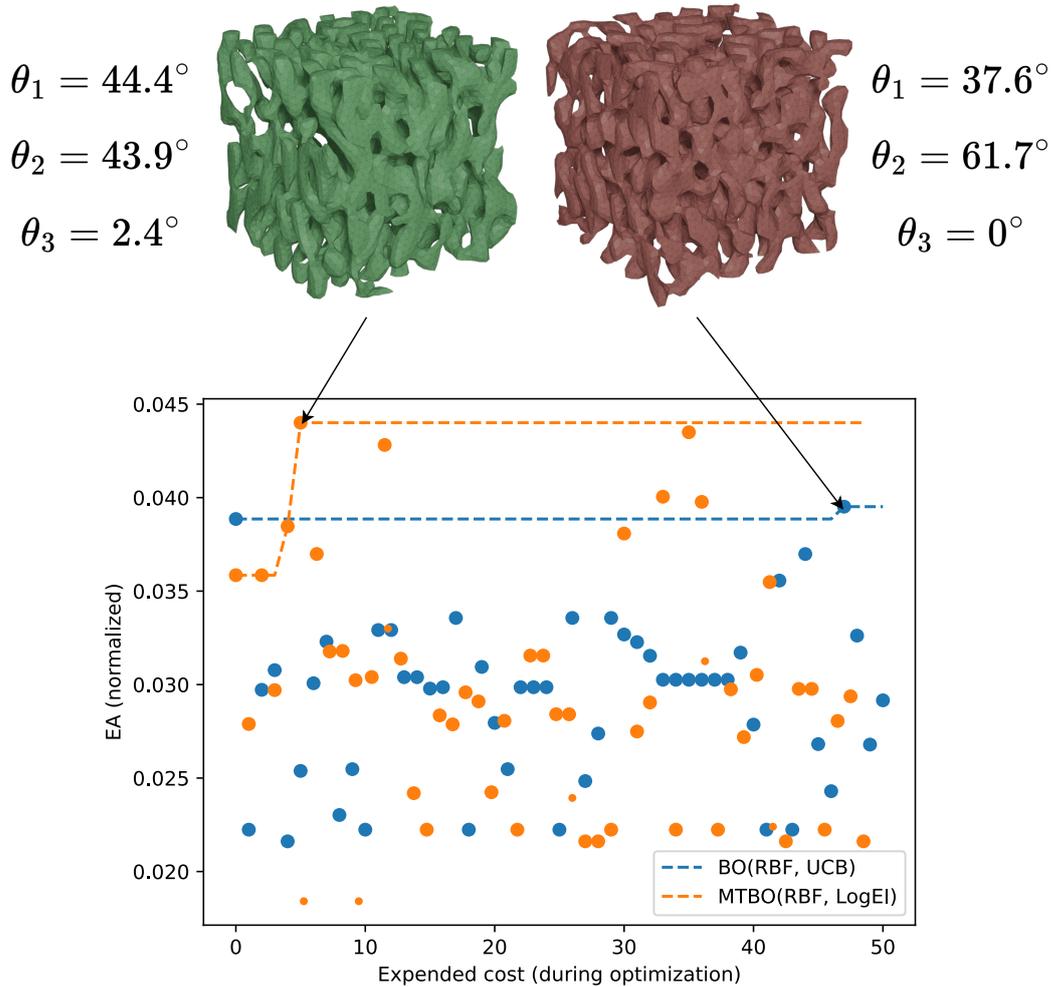}
		\caption{The EA-optimal spinodoid structures and $\theta$-values (Tables \ref{tab:spinodal-EA-3D-opt-rec} and \ref{tab:spinodal-EA-3D-mf-opt-rec}) found by the best-performing BO (top right) and the best MTBO (top left) schemes, superimposed with the optimization histories. Low-fidelity EA evaluations during the MTBO run are indicated with the smaller orange dots.}
		\label{fig:spinodal-EA-3D-mf-opt-vis2}
	\end{figure}

	In all, this means that the multi-task Bayesian optimization methodology is more reliable to locate a better optimum when it has access to faster simulations with lower fidelity. As a summary, Figure \ref{fig:spinodal-EA-3D-mf-opt-vis2} presents a direct comparison between the histories of the best-performing BO algorithm and the best-performing MFBO algorithm.
	
	
	As can be seen from Figure \ref{fig:spinodal-EA-3D-mf-opt-vis2}, the MTBO with the RBF covariance kernel and the logarithmic EI acquisition is able to adaptively sample multiple designs with superior EA values throughout the optimization run, compared to the best-performing BO algorithm.
	
	\section{Conclusions}
	
	This paper focused on optimizing the energy absorption of spinodoid structures. To this end, a potential set of design parameters was decided upon. Axial projection and variance-based sensitivity analysis showed that the dimensionality of the design space could be reduced due to the strong and monotonous dependency of EA on the density $\rho$. Subsequently, on the reduced design space by fixing $\rho=0.3$, both BO with GPR as well as MTBO were applied to search for the design configuration with the highest energy absorption, given an initial DoE containing samples that have also been used for SSA. It was shown that, with the given computational budget constraints, that MTBO outperforms single-fidelity BO by 11\% in the best-case scenario. This is by virtue of transfer learning made possible by computationally cheaper samples.
	
	To conclude, a variety of potential extensions to this research is suggested.
	\begin{itemize}
		\item \textbf{Experimental validation}. The optimized results in this manuscript have been purely based on simulation results. Given the precedent with regards to additively manufactured spinodoid structures \cite{wu2023additively}, it is desirable to experimentally validate the optimized design parameters by means of a real-life loading test on a manufactured spinodoid sample.
		
		\item \textbf{Design and fidelity space considerations}. Following the most popular convention in multi-fidelity literature, a total of $M=2$ fidelity levels were assumed to be part of the multi-fidelity data structure. However, the effect of an increase in $M$ is yet to be studied in the context of the energy absorption maximization problem. For example, how do the MTBO results presented earlier compare to a scenario with an intermediate fidelity level ($M=3$)?
		
		Furthermore, the Sobol' sensitivity analysis on the 4D case has showcased a significant skew towards reliance on the relative density of the material. While this allowed for the simplification of the design problem at hand, it will be especially interesting to apply MFBO to a design problem with a higher design space dimensionality. One possibility is to consider graded spinodoids \cite{liu2024mechanical}.
		
		\item \textbf{Multi-objective considerations}. While BO and MFBO have proven to optimize the energy absorption fairly well across the design space, it is not the only objective of interest in this class of material design problems. Minimizing the peak force (PF) that the material bears during loading is another important objective. One way to handle multiple objectives by means of a single-objective method is by scalarizing the EA and PF values into one objective \cite{marler2010weighted,knowles2006parego}. This scalarized objective can then be optimized by way of a single-fidelity or multi-fidelity BO method.
		
		Lastly, owing to the success of applying multi-objective Bayesian optimization to cooling duct design \cite{ellison2021robust} and a similar spinodoid design problem \cite{kansara2024multi}, it has been of interest to combine multi-fidelity and multi-objective BO (MOBO) into one framework. Previously, literature has showcase that it is possible to translate an MFBO problem into an MOBO problem by introducing a trust function \cite{irshad2024leveraging}. Heuristically speaking, a trust function measures the level of validity to use a low-fidelity objective to model a high-fidelity objective, similar to the usage of Pearson's correlation coefficient in Figure \ref{fig:spinodal-EA-3D-interfidelity-EA}.
	\end{itemize}
	
	\section*{Data availability}
	All presented data and the implemented workflow presented in this manuscript are open-source and accessible via GitHub: \url{https://github.com/llguo95/MFB}
	
	\section*{Acknowledgement}
	This work was supported by the ECSEL Joint Undertaking (JU) under Grant 826417. The JU receives support from the European Union’s Horizon 2020 research and innovation program and Germany, Austria, Spain, Finland, Hungary, Slovakia, Netherlands, Switzerland.
	
	Wei Tan, Hirak Kansara and Siamak F. Khosroshahi furthermore acknowledges the financial support from the EPSRC New Investigator
	Award (grant No. EP/V049259/1).
	
	Lastly, the authors extend much gratitude to Dr. Miguel A. Bessa for the valuable discussions, insights and support throughout the project leading to this manuscript.
	
	\bibliographystyle{elsarticle-num}
	\bibliography{dissertation.bib}
		
	\end{document}